\documentclass[onefignum,onetabnum]{siamonline220329}
\usepackage{graphicx}
\ifpdf
  \DeclareGraphicsExtensions{.eps,.pdf,.png,.jpg}
\else
  \DeclareGraphicsExtensions{.eps}
\fi
\usepackage{enumitem}
\setlist[enumerate]{leftmargin=.3in}
\setlist[itemize]{leftmargin=.3in}
\usepackage{pgfplots}
\usepackage{mathtools}
\usepackage{amsmath,floatflt,amsfonts,amscd,graphics} \usepackage{latexsym,mathrsfs}
\usepackage{verbatim}
\usepackage{algorithm}
\usepackage{algorithmic}
\usepackage{url}
\usepackage{color}
\usepackage{hyperref}
\usepackage{booktabs}
\usepackage{subfigure,tabularx}
\usepackage{adjustbox,caption}
\usepackage{amsopn}
\usepackage{hypcap,float}
\usepackage{pifont}
\newsiamthm{remark}{Remark}

\DeclareMathOperator{\domain}{dom} 
\DeclareMathOperator{\interior}{int} 

\DeclareMathOperator{\argmin}{argmin}

\newcommand{\mbbR}{\mathbb{R}}
\newcommand{\mcalX}{\mathcal{X}}
\newcommand{\mcalP}{\mathcal{P}}
\newcommand{\tildeW}{\tilde{W}}
\newcommand{\tildeH}{\tilde{H}}
\newcommand{\tildex}{\tilde{x}}
\newcommand{\mcalB}{\mathcal{B}}
\newcommand{\mcalD}{\mathcal{D}}
\newcommand{\mcalL}{\mathcal{L}}
\newcommand{\bbE}{\mathbb{E}}
\newcommand{\bbR}{\mathbb{R}}
\newcommand{\barphi}{\bar{\phi}}
\newcommand{\mcalY}{\mathcal{Y}}
\newcommand{\hatH}{\hat{H}}
\newcommand{\hatW}{\hat{W}}

\headers{Block Majorization Minimization with Extrapolation}{L.T.K. Hien, V. Leplat, and N. Gillis}

\title{Block Majorization Minimization with Extrapolation and Application to $\beta$-NMF 
\thanks{NG and LTKH acknowledge the support by the European Union (ERC consolidator, eLinoR, no 101085607).}}

\author{
Le Thi Khanh Hien\thanks{Acapela Group, 
Boulevard Dolez 33, 
7000 Mons, Belgium  
  (\email{khanhhiennt@gmail.com}).}
 \and 
   Valentin Leplat\thanks{Institute of
data sciences, Faculty of Engineering, 
Innopolis Univerity, Innopolis, Russia
  (\email{ V.Leplat@innopolis.ru}).}  
\and Nicolas Gillis\thanks{Department of Mathematics and Operational Research, University of Mons, Mons, Belgium
  (\email{nicolas.gillis@umons.ac.be}).}
  }

\definecolor{brightpink}{rgb}{1.0, 0.0, 0.5} 

\newcommand{\ngir}[1]{{{\color{black} #1}}}         

\newcommand{\ngi}[1]{{{\color{black} #1}}}   

\newcommand{\ngiii}[1]{{{\color{black} #1}}}

\newcommand{\Vali}[1]{{{\color{black} #1}}}                 
\newcommand{\revise}[1]{{{\color{black} #1}}}                 

\newcommand{\revision}[1]{{{\color{black} #1}}}                 

\definecolor{brightpink}{rgb}{1.0, 0.0, 0.5}

\begin{document}
\maketitle
\begin{abstract}
We propose a Block Majorization Minimization method with Extrapolation (BMMe) for solving a class of multi-convex optimization problems. The extrapolation parameters of BMMe are updated using a novel adaptive update rule. By showing that block majorization minimization can be reformulated as a block mirror descent method, with the Bregman divergence adaptively updated at each iteration, we establish subsequential convergence for BMMe. We use this method to design efficient algorithms to tackle nonnegative matrix factorization problems with the $\beta$-divergences ($\beta$-NMF) for $\beta\in [1,2]$. These algorithms, which are multiplicative updates with extrapolation, benefit from our novel results that offer convergence guarantees. We also empirically illustrate the significant acceleration of BMMe for $\beta$-NMF through extensive experiments.
\end{abstract} 


\section{Introduction} In this paper, we consider the following class of multi-convex  optimization problems: 
\begin{equation}
    \label{eq:compositev2}
    \min_{x_i\in\mcalX_i} f(x_1,\ldots,x_s),
\end{equation}
where $x=(x_1,\ldots,x_s)$ is decomposed into $s$ blocks, \revise{$\mcalX_i\subseteq\mathbb E_i$} is a closed convex set for $i=1,\ldots,s$, 
 \revise{$\mathbb E_i$ is a finite
dimensional real linear space equipped with the norm $\|\cdot\|_{(i)}$ and the inner product $\langle \cdot,\cdot\rangle_{(i)}$ (we will omit the lower-script $(i)$ when it is clear in the context)}, $\mcalX=\mcalX_1\times\ldots\mcalX_s\subseteq\interior\domain (f)$, \revise{$f:\mathbb E=\mathbb E_1\times\ldots\times\mathbb E_s\to\mbbR\cup\{+\infty\}$} is a differentiable function over the interior of its domain. \revise{Throughout the paper, we assume $f$ is lower bounded and  multi-convex}, that is, $x_i\mapsto f(x)$ is convex. 

\subsection{Application to $\beta$-NMF, $\beta\in [1,2]$}  \label{sec:applibetaNMF}

Nonnegative matrix factorization (NMF) is a standard linear dimensionality reduction method \Vali{tailored} for data sets with nonnegative values~\cite{Lee99}. \ngi{Given a nonnegative data matrix, $X \geq 0$, and a factorization rank, $r$, NMF aims to find two nonnegative matrices, $W$ with $r$ columns and $H$ with $r$ rows, such that $X \approx WH$.} The $\beta$ divergence is a widely used objective function in NMF to measure the difference between the input matrix, $X$, and its low-rank approximation, $WH$~\cite{Fevotte2011}. 
This problem is referred to as $\beta$-NMF \ngir{and can be formulated in the form of~\eqref{eq:compositev2} with two blocks of variables, $W$ and $H$, as follows:} given $X \in\mbbR^{m\times n}_+$ and $r$, solve  
\begin{equation}
    \label{betaNMF}
    \min_{\substack{W\in\mbbR^{m\times r},W\geq\varepsilon, \\ H\in\mbbR^{r\times n}, H\geq\varepsilon}} D_{\beta}(X,WH) ,
\end{equation}
where 
$D_{\beta}(X,WH)=\sum_{i=1}^m\sum_{j=1}^n d_\beta(X_{ij}, (WH)_{ij})$, with  
\begin{align*}
d_{\beta}(x,y) =  \left\{ 
\begin{array}{cc}
 x \log \frac{x}{y} -x + y & \text{for } \beta = 1, \\ 
\frac{1}{\beta (\beta-1)} \left(x^\beta + (\beta-1)y^\beta - \beta xy^{\beta-1}\right) &  \text{for } 1<\beta \leq 2. 
 \end{array}
\right. 
\end{align*} 
When $\beta = 2$, 
$d_\beta$ is the Euclidean distance, and when $\beta=1$ the Kullback-Leibler (KL) divergence; see Section~\ref{sec:KLnmf} for a discussion on the KL divergence. 
 Note that we consider a small positive lower bound, $\varepsilon>0$, for $W$ and $H$ to allow the convergence analysis. In practice, we use the machine epsilon for $\varepsilon$, which does not influence
the objective function much~\cite{HienNicolasKLNMF}.


\subsection{Previous works}  \label{sec:prevworks}

Block coordinate descent (BCD) methods serve as conventional techniques for addressing the multi-block Problem~\eqref{eq:compositev2}. These approaches update one block of variables at a time, while keeping the values of the other blocks fixed. There are three main types of BCD methods: classical BCD \cite{GRIPPO20001,Tseng2001}, proximal BCD \cite{GRIPPO20001}, and proximal gradient BCD \cite{Beck2013,Bolte2014,Tseng2009}. These methods fall under the broader framework known as the block successive upper-bound minimization algorithm (BSUM), as introduced in \cite{Razaviyayn2013}. In BSUM, a block $x_i$ of $x$ is updated by minimizing a majorizer (also known as an upper-bound approximation function, or a surrogate function; see Defintion~\ref{def:majorizer}) of the corresponding block objective function.



To accelerate the convergence of BCD methods for nonconvex problems, a well-established technique involves the use of extrapolation points in each block update, as seen in 
 \cite{Xu2013,Ochs2014,Pock2016,Ochs2019,HienICML2020}. Recently, \ngir{\cite{HPGTITAN}} proposed TITAN, an inertial block majorization-minimization framework for solving a more general class of multi-block  composite optimization problems than \eqref{eq:compositev2}, \revise{in which $f$ is not required to be multi-convex}. TITAN updates one block of $x$ at a time by selecting a majorizer function for the corresponding block objective function, incorporating inertial force into this majorizer, and then minimizing the resulting inertial majorizer. Through suitable choices of majorizers and extrapolation operators, TITAN recovers several known inertial methods and introduces new ones, as detailed in \cite[Section 4]{HPGTITAN}. TITAN has proven highly effective in addressing low-rank factorization problems using the Frobenius norm, as demonstrated in \cite{HieniADMM,HPGTITAN,HienICML2020,Vu-Thanh2021}. 
\ngi{However, to ensure convergence, TITAN requires the so-called nearly sufficiently decreasing property (NSDP) of the objective function between iterations. The NSDP is satisfied in particular when the majorizer is strongly \Vali{convex} or when the error function, that is, the difference between the majorizer and the objective, is lower bounded by a quadratic function~\cite[Section 2.2]{HPGTITAN}.} 
Such requirements \Vali{pose} issues in some situations; for example the Jensen surrogate used to design the multiplicative updates (MU) for standard $\beta$-NMF (see Section~\ref{alg:MUe-beta} for the details) lacks strong convexity, nor the corresponding error function is lower bounded by a quadratic function. \revise{In other words, although TITAN does not require $f$ to be multi-convex, utilizing TITAN for accelerating the MU in the context of $\beta$-NMF 
is very challenging. This scenario corresponds to a specific instance of Problem \eqref{eq:compositev2}.}

\revise{Consider \ngir{$\beta$-NMF}~\eqref{betaNMF}.} For $\beta = 2$, NMF admits very efficient BCD algorithms with \Vali{theoretically grounded} extrapolation~\cite{HienICML2020} \Vali{and heuristic-based extrapolation mechanisms~\cite{Ang2018}}. 
Otherwise, the most widely used algorithm to tackle $\beta$-NMF are the multiplicative updates (MU): \revision{for $1 \leq \beta \leq 2$,}   
 \begin{equation} \label{eq:standardMU}  
H  
\leftarrow 
\text{MU}(X,W,H) = 
\max\left(\varepsilon ,  
H  \circ 
\frac{
\big[ W^\top \frac{[X]}{[WH]^{.(2-\beta)}}  \big]
}
{
\big[  W^\top [WH]^{.(\beta-1)}]  \big] 
}
\right), 
\end{equation} 
and $W^\top   
\leftarrow \text{MU}\big(X^\top,H^\top,W^\top\big)$, 
 where  $\circ$ and $\frac{[\cdot]}{[\cdot]}$ are the component-wise product and division between two matrices, respectively, and \ngir{$(.)^{.x}$ denotes the component-wise exponent}.  
The MU are guaranteed to decrease the objective function~\cite{Fevotte2011}; 
see Section~\ref{sec:prel-MU} for more details.  
Note that, by symmetry of the problem,  since $X = WH \iff X^\top = H^\top W^\top$,  the MU for $H$ and $W$ are the same, up to transposition. 

\Vali{As far as we know, there is currently no existing algorithm in the literature that accelerates the MU while providing convergence guarantees.}  \revise{On the other hand,} it is worth noting that an algorithm with a guaranteed convergence in theory may not always translate to practical success. 
 For instance,  the block mirror descent method, while being the sole algorithm to ensure global convergence in KL-NMF, does not yield effective performance in real applications, as reported in \cite{HienNicolasKLNMF}.
 
\subsection{Contribution and outline of the paper}

Drawing inspiration from the versatility of the BSUM \Vali{framework~\cite{Razaviyayn2013}} and the acceleration effect observed in TITAN~\Vali{\cite{HPGTITAN}}, we introduce BMMe, which stands for Block Majorization Minimization with Extrapolation, to address Problem~\eqref{eq:compositev2}. \revise{Leveraging the multi-convex structure in Problem~\eqref{eq:compositev2}, BMMe does not need the NSDP condition to ensure convergence; instead, block majorization minimization for \ngir{the} multi-convex Problem~\eqref{eq:compositev2} is reformulated as a block mirror descent method, wherein the Bregman divergence is adaptively updated at each iteration, and the extrapolation parameters in BMMe are dynamically updated using a novel adaptive rule.  
We establish subsequential convergence for BMMe, apply BMMe to 
tackle $\beta$-NMF problems with $\beta\in [1,2]$, and showcase the \Vali{obtained} acceleration effects through extensive \Vali{numerical }experiments.} 

\begin{figure}[ht!]
\begin{center}
\begin{tabular}{c}
\includegraphics[width=0.55\textwidth]{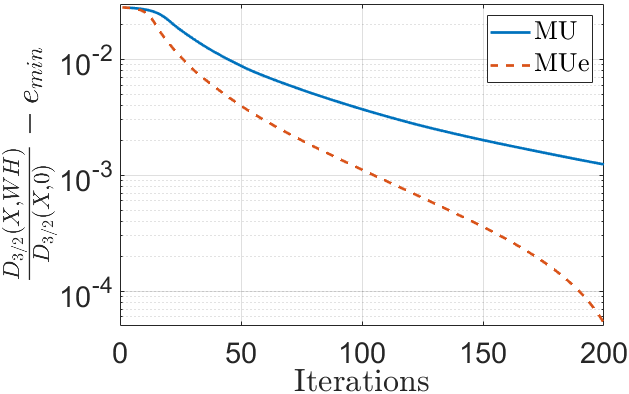} 
\end{tabular}
\caption{MU vs.\ MUe with Nesterov extrapolation sequence~\eqref{eq:exrapolNes} on the CBCL data set with $r=49$. Evolution of the median relative  objective function values, over 10 random initial initializations, of $\beta$-NMF for $\beta = 3/2$ minus the smallest relative objective  function found among all runs (denoted $e_{min}$).}  
\label{fig:cbcl}
\end{center}
\end{figure}

\ngir{To give an idea of the simplicity and acceleration of BMMe, let us show how it works for $\beta$-NMF.} 
 \ngi{Let $(W,H)$ and $(W^p,H^p)$ be the current and previous iterates, 
respectively.  
 BMMe will provide the following  
 \ngir{MU with extrapolation} (MUe)}: 
\begin{align} 
\hat H  = H + \alpha_H [H -  H^p]_+,  
& \quad  
 H \leftarrow \text{MU}(X,W,\revise{\hat H}), \label{eq:MUeH}   
\end{align} 
and similarly for $W$.  
We will show that MUe not only allows us to empirically accelerate the convergence of the MU significantly for a negligible additional cost per iteration \revision{(see Remark~\ref{rem:timevsiter} below)} and a \Vali{slight} modification of the original MU,  
but has convergence guarantees (Theorem~\ref{thrm:BMMe}). 
We will discuss in details how to choose the extrapolation parameters $\alpha_W$ and $\alpha_H$ in Section~\ref{sec:BMMe}. 
\Vali{It is important to note that no restarting step is required to ensure convergence. As a result, there is no \ngir{need} to compute objective function values during the iterative process, which would otherwise incur significant computational expenses.}
We will also show how to extend the MUe to regularized and constrained $\beta$-NMF \Vali{problems} in Section~\ref{sec:KLnmf}. 
Figure~\ref{fig:cbcl} illustrates the behavior of MU vs.\ MUe on the widely used CBCL facial image data set with $r=49$, as in the seminal paper \ngir{of~\cite{Lee99} who introduced NMF},  
and with 
$\beta = 3/2$. 
MUe is more than twice faster than MU: \ngir{over 10 random initializations,} 
it takes MUe between 
88 and 95 iterations with a median of 93 to obtain an objective smaller than the MU with 200 iterations. 
We will provide more experiments in Section~\ref{sec:experiment} that \Vali{confirm} the significant acceleration effect of MUe. 

\revision{
\begin{remark}[Time vs.\ Iterations] \label{rem:timevsiter} 
The extra cost of MUe compared to MU is only the computation of the extrapolated point, $\hat H  = H + \alpha_H [H -  H^p]_+$. For the update of $H$, this costs $O(nr)$ operations and $O(nr)$ memory. 
    The MU itself requires the computation of $WH$ in $O(mnr)$ operations and $O(mn)$ memory, and multiplying $[X]./\big[[WH]^{.(2-\beta)}\big]$ and $[WH]^{.(\beta-1)}$ by $W^\top$ requires $O(mnr)$ operations. The same observation holds for $W$ where the role of $m$ and $n$ are exchanged. 
    For example, for the CBCL data set experiment in Figure~\ref{fig:cbcl}, with $m=361$, $n=2429$, $r=49$, MUe requires less than 1\% more time than MU: on 30 runs with 1000 iterations, the average time for MU is 11.63~s and for MUe it is 11.71~s, which is about 0.7\% more than MU.   
    Given this negligible difference, for simplicity we report the iteration number instead of the computational time when comparing MU with MUe. When comparing with other algorithms, we will use the computational time.  
\end{remark}
}

The paper is organized as follows. 
In the next section, we provide preliminaries on majorizer functions, the majorization-minimization method, the multiplicative updates for $\beta$-NMF, and nonconvex optimizations.  
In Section~\ref{sec:BMMe}, we describe our new proposed method, BMMe, 
and prove its convergence properties. 
In Section~\ref{sec:betanmf}, we apply BMMe to solve standard $\beta$-NMF, as well as  an important regularized and constrained KL-NMF model, namely the minimum-volume KL-NMF. 
We report numerical results in Section~\ref{sec:experiment}, and conclude the paper in Section~\ref{sec:conclusion}.

\paragraph{Notation} We denote $[s]=\{1,\ldots,s\}$. 
We use $I_{\mcalX}$ to denote the indicator function \Vali{associated to the set} $\mcalX$. 
For a given matrix $X$, we denote by $X_{:j}$ and $X_{i:}$ the $j$-th column and the $i$-th row of $X$, respectively. We denote the nonnegative part of $X$ as $[X]_+ =  \max(0,X)$ where the max is taken component wise. Given a multiblock differentiable function $f: x=(x_1,\ldots,x_s)\in\mathbb E\mapsto f(x)$, we use $\nabla_i f(x)$ to denote its partial derivative $\frac{\partial f(x)}{\partial x_i}$. We denote \Vali{by} $e$ the vector of all ones of appropriate dimension.

\section{Preliminaries}


\label{sec:prel}
\subsection{Majorizer, majorization-minimization method, and application to $\beta$-NMF} 
\label{sec:prel-MU} 

We adopt the following definition for a majorizer. 

\begin{definition}[Majorizer] \label{def:majorizer}
    A continuous function $g:\mcalY\times\mcalY\to\mbbR$ is called a majorizer (or a surrogate function) of a differentiable function $f$ over $\mcalY$ if the following conditions are satisfied: 
   \begin{itemize}
    \item[(a)]  $g(x,x)=f(x)$ for all $x\in\mcalY$,
    
 \item[(b)]  $g(x,y)\geq f(x)$ for all $x,y\in\mcalY$, and

\item[(c)]  $\nabla_1 g(x,x)=\nabla f(x)$ for all $x\in\mcalY$. 
    \end{itemize}
\end{definition}

\revise{It is important noting that Condition (c) can be replaced by the condition on directional derivatives as in \cite[Assumption 1 (A3)]{Razaviyayn2013}, and the upcoming analysis still holds. For simplicity, we use Condition~(c) in this paper.} 
Let us give some examples of majorizers. 
 The second and third one will play a pivotal role in this paper. \Vali{More examples of majorizers can be found  in~\cite{Mairal_ICML13,SunMM2017,HPGTITAN}.} 
 
\noindent \emph{1. Lipschitz gradient majorizer} (see, e.g.,  \cite{Xu2013}). 
    If $\nabla f$ is $L$-Lipschitz continuous over $\mcalY$, then 
    \begin{equation*}\label{eq:LGD_majo}
        g(x,y)=f(y)+\langle \nabla f(y),x-y \rangle + \frac{L}{2} \|x-y\|^2
    \end{equation*}
    is called \ngi{the} Lipschitz gradient majorizer of $f$. 
    
    
    \noindent \emph{2. Bregman majorizer} (see, e.g., \cite{melo2017}). 
    Suppose there exists a differentiable convex function $\kappa$ 
    \ngir{and $L > 0$} such that $x \mapsto L \kappa(x) - f(x)$ is convex. Then 
    \begin{equation} \label{eq:BMajo} 
        g(x,y)  = f(y) + \langle \nabla f(y), x-y \rangle  
          + L\big(\kappa(x)-\kappa(y)-\langle \nabla \kappa(y),x-y \rangle\big)   
    \end{equation} 
    is called a Bregman majorizer of $f$ with kernel function~$\kappa$. \revise{When $\kappa=\frac{1}{2} \| .\|^2$, the Breman majorizer coincides with the Lipschitz gradient majorizer.} 
    
    
    \noindent \emph{3. Jensen majorizer} (see, e.g., \cite{Dempster1977,Neal1998,Lange2000}). 
    Suppose $\tilde f:\mbbR\to\mbbR$ is a convex function and $\omega\in \mbbR^r$ is a given vector. Define $
    f:x\in\mbbR^r\mapsto \tilde f(\omega^\top x)$. 
    Then 
    $$
    g(x,y)=\sum_{i=1}^r \alpha_i \tilde f\left(\frac{\omega_i}{\alpha_i}(x_i-y_i) + \omega^\top y\right), 
    $$ 
    where $\alpha_i\geq 0$, $\sum_{i=1}^r \alpha_i=1$, and $\alpha_i\ne0$ whenever $\omega_i\ne0$, is called a Jensen majorizer of $f$. The term ``Jensen" in the name of the majorizer \Vali{comes from the fact that} the Jensen inequality for convex functions is used to form the majorizer. Indeed, \ngir{by the Jensen inequality}, 
    $$
       \sum_{i=1} ^r \alpha_i \tilde f\left(\frac{\omega_i}{\alpha_i}(x_i-y_i) + \omega^\top y\right) 
       \; \geq \; \tilde f \left(\sum_{i=1} ^r \alpha_i \left[\frac{\omega_i}{\alpha_i}(x_i-y_i) + \omega^\top y \right]\right)= \tilde f\big(\omega^\top x\big).
   $$
    Choosing $\alpha_i = \frac{\omega_i y_i}{\omega^\top y}$, $g(x,y)=\sum_{i=1} ^r \frac{\omega_i y_i}{\omega^\top y} \tilde f\big( \frac{\omega^\top y}{y_i} x_i \big)$ is an example of a Jensen surrogate of $f$, if $g$ is well-defined.  
    
    

Given a majorizer of $f$, \Vali{the minimization of} $f$ over $\mcalX$ can be \Vali{achieved} by iteratively minimizing its majorizer, using 
$$
x^{t+1} \; \in \; \underset{x\in\mcalX}{\argmin} \, g(x,x^t),
$$ 
where $x^t$ denotes the $t$-th iterate. 
This is the majorization-minimization (MM) method 
\ngi{which guarantees, by properties of the majorizer, that $f(x^{t+1}) \leq f(x^{t})$ for all $t$; 
see \cite{Hunter2004, SunMM2017} for tutorials}.

\paragraph{Example with the Multiplicative Updates for $\beta$-NMF} 


\ngi{
The standard MU for $\beta$-NMF, given in~\eqref{eq:standardMU}, can be derived using the MM method. By symmetry of the problem, let us focus on the update of $H$. Moreover, we have that $D_\beta(X,WH) = \sum_i D_\beta(X_{:i},W H_{:i})$, \Vali{that is, the objective function is separable w.r.t. each column of $H$}, and hence one can focus w.l.o.g. on the update of a single column of $H$. 
Let us thefore provide a majorizer for $D_\beta(v,W h)$, and show how its closed-form solution leads to the MU~\eqref{eq:standardMU}. 
}

The following proposition, \revise{which is a corollary of \cite[Theorem 1]{Fevotte2011}}, provides a Jensen majorizer for  
$h\in\mbbR^r\mapsto D_{\beta}(v,Wh):=\sum_{i=1}^m d_\beta(v_{i},(Wh)_{i})$ with $\beta\in[1,2]$, where $v\in\mbbR^m_+$ and $W\in\mbbR^{m\times r}_+$ are given. 
\begin{proposition}[\cite{Fevotte2011}]
\label{prop:majorizer}
    Let us denote $\tilde{v} = W\tilde{h}$, and let $\tilde{h}$ be such that $\tilde{v}_{i}>0$ and $\tilde{h}_{i}>0$ for all $i$. 
    Then the following function is a majorizer for $h\mapsto D_{\beta}(v,Wh)$ with $\beta\in[1,2]$: 
    \begin{equation} 
\label{betaNMF-majorizer}
g(h,\tilde{h})=\sum_{i=1}^m\sum_{k=1}^r\frac{W_{ik}\tilde{h}_{k}}{\tilde{v}_{i}} d_{\beta}\left(v_i, \tilde v_i \frac{h_k}{\tilde h_k} \right) . 
\end{equation}
\end{proposition}

With this surrogate, the MU obtained via the MM method~\cite[Eq.~4.1]{Fevotte2011} 
are \Vali{as follows}:  
\begin{align*}  
 \argmin_{x \geq \varepsilon}  g(x , \tilde{h})
 \; = \; 
 \max\left(\varepsilon,  \tilde{h} \circ \frac{\big[W^\top \frac{[v]}{[W\tilde{h}]^{.(2-\beta)}}\big]}{\big[W^\top [W\tilde{h}]^{.(\beta-1)} \big]} \right), 
 \end{align*} 
 which leads to the MU in the matrix form~\eqref{eq:standardMU}.  
The term \textit{multiplicative} in the name of the algorithm is because 
\Vali{the new iterate is obtained by an element-wise multiplication between the current iterate, $\tilde{h}$, and a correction factor.}
 

\subsection{Critical point and coordinate-wise minimizer} 
 \label{sec:prelnnopt}

{Let us define three key notions for our purpose: subdifferential, critical point and coordinate-wise minimizer.}  
  
\begin{definition}[Subdifferentials] Let $g: \bbE\to \bbR\cup \{+\infty\} $ be a proper lower semicontinuous function.  
\label{def:dd}
\begin{itemize}
\item[(i)] For each $x\in{\rm dom}\,g,$ we denote $\hat{\partial}g(x)$ as
the Fr\'echet subdifferential of $g$ at $x$ which contains vectors
$v\in\mathbb{E}$ satisfying 
\begin{align*}
\liminf_{y\ne x,y\to x}\frac{1}{\left\Vert y-x\right\Vert }\left(g(y)-g(x)-\left\langle v,y-x\right\rangle \right)\geq 0.
\end{align*}
If $x\not\in{\rm dom}\:g,$ then we set $\hat{\partial}g(x)=\emptyset.$
\item[(ii)] The limiting-subdifferential $\partial g(x)$ of $g$ at $x\in{\rm dom}\:g$
is defined as follows: 
\begin{align*}
\partial g(x) := \big\{ v\in\mathbb{E}:  \; \exists x^{k}\to x,\,g\big(x^{k}\big)\to g(x),  
    v^{k}\in\hat{\partial}g\big(x^{k}\big),\,v^{k}\to v \big\} .
\end{align*}
\end{itemize}
Partial subdifferentials with respect to a subset of the variables are defined analogously by considering the other variables as parameters.
\end{definition}

\begin{definition}[Critical point] 
\label{def:type2}
We call $x^{*}\in \rm{dom}\,g$ a critical point of $g$ if $0\in\partial g\left(x^{*}\right).$ 
\end{definition}
If $x^{*}$ is a local minimizer of $g$ then it is a critical point of $g$. 

\begin{definition}[Coordinate-wise minimizer]
    We call $x^*\in\domain f$ a coordinate-wise minimizer of Problem~\eqref{eq:compositev2} if 
    \begin{align*}
      f(x_1^*,\ldots,x_{i-1}^*,x_i^*, x_{i+1}^*,\ldots,x_s^*)  
       \leq f(x_1^*,\ldots,x_{i-1}^*,x_i, x_{i+1}^*,\ldots,x_s^*), \forall\, x_i \in\mcalX_i,  
    \end{align*}
    or, equivalently, 
    $
        \langle \nabla_{i} f(x^*), x_i - x_i^* \rangle \geq 0, \forall x_i \in\mcalX_i. 
    $
\end{definition}
For Problem~\eqref{eq:compositev2}, a critical point  of $f(x)+\sum_{i=1}^s I_{\mcalX_i}(x_i)$,  must be a coordinate-wise minimizer.

\section{Block Majorization Minimization with Extrapolation (BMMe)}
\label{sec:BMMe}


\ngi{
In this section, we introduce BMMe, see Algorithm~\ref{alg:BMMe}, and then prove its convergence, see Theorem~\ref{thrm:BMMe}. }

\begin{algorithm}[ht!]
\caption{BMMe for solving Problem~\eqref{eq:compositev2}} \label{alg:BMMe}
\begin{algorithmic}[1]
\STATE Choose initial points $x^{-1}, x^0\in \domain (f)$. \ngi{(Typically, $x^{-1} =  x^0$.)} 
\FOR{$t=\ngi{0},\ldots$}
\FOR{$i=1,\ldots,s$}

\STATE \ngi{Extrapolate block $i$: 
\begin{equation} \label{eq:extrapolBMMe}
 \hat x_i^t \; = \; x_i^t \, +  \,  \alpha_i^t  \, \mathcal P_i \left(x_i^t-x_i^{t-1} \right), \quad \text{ where } 
\end{equation}
 $\bullet$ $\mathcal P_i$ is a 
 mapping such that \revise{$\hat x_i^t\in \interior\domain(f_i^t)$}, where \revise{$f_i^t$ is defined in \eqref{fit}} 
 (for example, 
 in $\beta$-NMF, we use $\mathcal P(a)=[a]_+$, see Section \ref{sec:betanmf}), 

 $\bullet$  $\alpha_i^t$ are the extrapolation parameters; 
 see Section~\ref{sec:convBMMe} for the conditions they need to satisfy.  
}
 
\STATE Update block $i$: 
\begin{equation}
\label{eq:BMDstep}
x^{t+1}_i = \argmin_{x_i\in \mathcal X_i} G^t_i(x_i,\hat x_i^t),
\end{equation}
\revise{where $G^t_i$ is a majorizer of $f_i^t$ over its domain.}
\ENDFOR
\ENDFOR
\end{algorithmic}
\end{algorithm} 

\revision{
\begin{remark}[Mappings $\mathcal P_i$] BMMe requires the mappings 
$\mathcal P_i$, one for each block of variables. 
There are multiple choices possible for a given case.  
For example, if $\domain(f_i^t)$ is the full space, then both $[.]_+$ and the identity mapping would satisfy condition $\hat x_i^t\in \interior\domain(f_i^t)$. 
If $\domain(f_i^t)$ is a convex cone, then $\mathcal P_i \left(x_i^t-x_i^{t-1} \right) = [ x_i^t - x_i^{t-1}  ]_{\mathcal C}$ would satisfy the condition, where $[\cdot]_{\mathcal C}$ denotes the projection onto any  closed convex subset $\mathcal C$ of $\domain(f_i^t)$. This is what we use for $\beta$-NMF, with $\mathcal P_i(a)=[a]_+$ for all $i$. 
\end{remark}
} 
 
\subsection{Description of BMMe}

BMMe, \revise{see Algorithm~\ref{alg:BMMe}}, updates one block of variables at a time, say $x_i$, by minimizing a majorizer \revise{$G^t_i$ of} 
\begin{equation}
\label{fit}
    \revise{f_i^t(x_i)  := f\big( x_1^{t+1},\ldots,x_{i-1}^{t+1},x_i,x_{i+1}^t,\ldots,x_s^t\big), }
\end{equation} 
where the other blocks of variables $\{x_j\}_{j \neq i}$ are fixed, and $t$ is the iteration index. The main difference \revise{between BMMe and}  standard block MM \revise{(BMM)} is that the majorizer \revise{in BMMe is evaluated} \Vali{at}  the extrapolated block, $\hat x_i^t$ given in~\eqref{eq:extrapolBMMe}, \revise{while it is evaluated at the previous iterate  $x_i^t$ in BMM.}
The MUe~\eqref{eq:MUeH} described in Section~\ref{sec:applibetaNMF} 
follow exactly this scheme; we elaborate more on this specific case in  Section~\ref{sec:beta-nmf}.

\ngi{
\revise{In the following, we explain the notation that will be used in the sequel}, and then state the convergence of BMMe. 
} 
\begin{itemize}
 \item   Denote $\bar f_i(x_i):= f(\bar{ x}_1,\ldots,\bar{ x}_{i-1},x_i,\bar{ x}_{i+1},\ldots,\bar{ x}_s)$, where $\bar x$ is fixed. As assumed, the function 
$
 \bar f_i(\cdot)$, for $i\in [s]$, 
is convex and admits a majorizer $G^{(\bar x)}_i(\cdot,\cdot)$ over its domain.

 \item Given $\bar x$ and $\tildex_i$, we denote $\xi_i^{(\bar x,\tildex_i)}(x_i)=G^{(\bar x)}_i(x_i,\tildex_i)$ (i.e., we fix $\bar x$ and $\tildex_i$) and 
\begin{align*}
\mcalD_{\bar x,\tildex_i}(x_i,x'_i) & = \mcalB_{\xi_i}(x_i,x'_i) \\ 
& =\xi_i(x_i)-\xi_i(x'_i)-\langle \nabla\xi_i(x'_i),x_i-x'_i\rangle,
\end{align*}
where we omit the upperscript of $\xi_i^{(\bar x,\tildex_i)}$ for notation succinctness. Using this notation, we can write
\begin{equation}
    \label{Bregman_majorizer}
G^{(\bar x)}_i(x_i,\tildex_i)
=\bar f_i(\tildex_i) + \langle \nabla \bar f_i(\tildex_i), x_i -\tildex_i\rangle + \mcalD_{\bar x,\tildex_i}(x_i,\tildex_i), 
\end{equation}
where we use the facts that $\bar f_i(\tildex_i)=G^{(\bar x)}_i(\tildex_i,\tildex_i)$ and $\nabla\bar f_i(\tildex_i)=\nabla_1 G^{(\bar x)}_i(\tildex_i,\tildex_i)$. 

 \item  Denote $
x^{(t,i)} =( x_1^{t+1},\ldots,x_{i-1}^{t+1},x^t_i,x_{i+1}^t,\ldots,x_s^t)$, 
 $G^t_i=G_i^{(x^{(t,i)})}$ be the majorizer of $f^t_i$\ngi{, which is the notation used in Algorithm~\ref{alg:BMMe}}, and $\mcalD^t_{\hat x_i}=\mcalD_{x^{(t,i)}, \hat x_i}$.   

\end{itemize}


\paragraph{Key observation for BMMe} Using the notation in~\eqref{Bregman_majorizer}, the MM update in~\eqref{eq:BMDstep} 
can be rewritten as 
\begin{equation}
    \label{BMD-2}
    x_i^{t+1} \in \underset{x_i\in\mcalX_i}{\argmin}f^t_i(\hat x_i) + \langle \nabla f^t_i(\hat x_i), x_i -\hat x_i\rangle + \mcalD^t_{\hat x_i}(x_i,\hat x_i),
\end{equation}
which has the form of a mirror descent step with the Bregman divergence  $\mcalD^t_{\hat x_i}(x_i,\hat x_i)$ being adaptively updated at each iteration.  
This observation will be instrumental in proving  the convergence of BMMe. 




\subsection{Convergence of BMMe} \label{sec:convBMMe}

We present the convergence of BMMe in Theorem~\ref{thrm:BMMe}. All the technical proofs, \revision{except for our main Theorem~\ref{thrm:BMMe}}, are relegated to Appendix~\ref{app:techproofs}.



\begin{theorem}
\label{thrm:BMMe}
     Consider BMMe described in Algorithm~\ref{alg:BMMe}  for solving Problem~\eqref{eq:compositev2}. \revise{We assume that the function $x_i\mapsto G^{(\bar x)}_i(x_i,\tildex_i)$ is convex for any given $\bar x$ and $\tildex_i$. 
     Furthermore, we assume the following conditions are satisfied. }
     \begin{itemize}
     
       \item[(C1)] \revise{\textbf{Continuity.} For $i\in [s]$, if $x^{(t,i)}\to \bar x$ when $t\to\infty$ then $G_i^t (\grave x^t_i,\acute x^t_i)\to G_i^{(\bar x)}(\grave x_i,\acute x_i)$ for any $\grave x^t_i\to \grave x_i$ and $\acute x^t_i \to \acute x_i$, and $G_i^{(\bar x)}(\cdot,\cdot)$ is a majorizer of $\bar f_i(\cdot)$.
         }
         
         \item[(C2)] \revise{\textbf{Implicit Lipschitz gradient majorizer.}}
At iteration $t$ of Algorithm~\ref{alg:BMMe},  
for $i\in [s]$,  \ngi{there exists a constant $C_i>0$ such that} 
\begin{equation} 
\label{Ci-upper}
\mathcal D^t_{\hat x_i^t}(x^t_i,\hat x_i^t) \; \leq \;  C_i \|x_i^t-\hat x_i^t\|^2 
\; = \;  C_i (\alpha_i^t)^2 \|\mcalP_i(x_i^t-x_i^{t-1})\|^2.
 \end{equation}
 
   \item[(C3)] \ngi{The sequence of extrapolation parameters satisfies}   
        \begin{equation}
            \label{converge_condition-iBMD}
        \sum_{t=1}^{\infty} (\alpha^t_{i})^{2}\big\|\mcalP_i(x_i^{t}-x_i^{t-1}) \big\|^{2}<+\infty,\mbox{for} \, \, i \in [s].
        \end{equation}
 \end{itemize}
 Then we have  
 \begin{equation}
 \label{quadratic-series}\sum_{t=1}^{\infty}\sum_{i=1}^s  \mathcal D^t_{\hat x_i^t}(x^t_i,x_i^{t+1}) < +\infty, 
 \end{equation}
 and \ngi{the sequence generated by BMMe (Algorithm~\ref{alg:BMMe})}, $\{x^t\}_{t\geq 0}$, is bounded if $f$ has bounded level sets. 
 
 Furthermore, under the following condition, 
 \begin{itemize}
     \item[(C4)] $\underset{t\to\infty}{\lim}\|x^t_i-x_i^{t+1}\| =0$ when $\underset{t\to\infty}{\lim} \mathcal D^t_{\hat x_i^t}(x^t_i,x_i^{t+1})=0$, 
 \end{itemize}  
  any limit point of $\{x^t\}_{t\geq 0}$ is a critical point of $f(x)+\sum_{i=1}^s I_{\mcalX_i}(x_i)$.    
\end{theorem}
\revision{Before proving Theorem~\ref{thrm:BMMe}, let us \ngi{discuss} its  conditions :}   
 \begin{itemize} 
  \item  If $\mcalD_{\bar x,\tildex_i}(x_i,\tildex_i) \leq C_i  \|x_i - \tildex_i\|^2$ then \eqref{Ci-upper}  is satisfied. This condition means $\bar f_i(\cdot)$ is actually upper bounded by a Lipschitz gradient majorizer. However, it is crucial to \ngi{realize} that the introduction of $C_i$ is primarily for the purpose of the convergence proof. Employing a Lipschitz gradient majorizer for updating $x_i$ is discouraged due to the potential issue of $C_i$ being excessively large (this situation can result in an overly diminishing/small step size, rendering the approach inefficient in practical applications). \revise{For example,  $C_i=O(1/\varepsilon^2)$ in the case of KL-NMF, see the proof of Theorem~\ref{thrm:MUe-kl}. }

 \item  If $G_i^t(\cdot,\hat x_i^t)$ is $\theta_i$-strongly convex  then Condition (C4) is satisfied (here $\theta_i$ is a constant independent of $\{x^{(t,i)}\}$ and $\{\hat x_i^t\}$), \ngi{since $\theta_i$-strongly convexity implies } 
 $$
 \mathcal D^t_{\hat x_i^t}(x^t_i,x_i^{t+1}) \geq \frac{\theta_i}{2} \|x^t_i-x_i^{t+1}\|^2. 
 $$
 
 \item \revise{We see that the update in~\eqref{BMD-2} has the form of an accelerated mirror descent (AMD) method \cite{Tseng2008} for one-block convex problem. Hence, it makes sense to involve the extrapolation sequences that are used in AMD. This strategy has been used in \cite{Xu2013,HienICML2020,HPGTITAN}.} An example of choosing the extrapolation parameters satisfying \eqref{converge_condition-iBMD} is 
    \begin{equation}
        \label{parameter}
        \begin{split}
   \alpha_{i}^t&=\min\left\{ \alpha^t_{Nes},\frac{c}{t^{q/2}}\frac{1}{\big\|\mcalP_i(x_i^{t}-x_i^{t-1}) \big\|}\right\},
   \end{split}
    \end{equation}
    where $q>1$, $c$ is any \ngi{large} constant and $\alpha^t_{Nes}$ is the extrapolation sequence defined by 
    $\eta_0=1$, 
    \begin{equation} \label{eq:exrapolNes}
    \eta_t=\frac12\Big( 1+\sqrt{1+4 \eta_{t-1}^2}\Big),
    \; \alpha^t_{Nes} = \frac{\eta_{t-1}-1}{\eta_t}. 
    \end{equation} 
    Another choice is replacing $\alpha^t_{Nes}$ in \eqref{parameter} by $\alpha^t_c=\frac{t-1}{t}$ \cite{Tseng2008}. 
          In our experiments, we observe that when $c$ is \ngi{large enough}, $\alpha_{i}^t$ coincides with $\alpha^t_{Nes}$ (or $\alpha^t_c$). 
         \ngiii{A motivation to choose $\alpha_{Nes}^t$ is that, {for one-block problem with an $L$-smooth convex objective}, BMMe with the {Lipschitz  gradient majorizer recovers the} famous Nesterov fast gradient method, which is an optimal first-order method. However, it is pertinent to note that the best choice may vary depending on the particular application and dataset characteristics.}  
          
    
 
    \end{itemize}

\revision{ 
\begin{proof}[Proof of Theorem~\ref{thrm:BMMe}] 
    Since $G^t_i(\cdot,\cdot)$ is a majorizer of $x_i\mapsto f^t_i(x_i)$, 
\begin{equation}
     \label{iBMD-ie1}
     f_i^t(x_i^{k+1}) \leq G^t_i(x_i^{k+1},\hat x^t_i)=f^t_i(\hat x^t_i) + \langle \nabla f^t_i(\hat x^t_i), x_i^{t+1} -\hat x^t_i\rangle + \mcalD^t_{\hat x^t_i}(x_i^{t+1},\hat x^t_i).
 \end{equation}
  Applying Proposition~\ref{prop:threepoint} in Appendix~\ref{sec:propthreepoint} for \eqref{BMD-2} with $\varphi(x_i)=  f^t_i(\hat x^t_i) + \langle \nabla f^t_i(\hat x^t_i), x_i -\hat x^t_i\rangle$ and fixing $z=\hat x^t_i$,  
  \begin{equation}
     \label{iBMD-ie2} 
     \varphi (x_i) + \mathcal D^t_{\hat x_i^t}(x_i,\hat x_i^t)\geq \varphi(x_i^{t+1}) + \mathcal D^t_{\hat x_i^t}(x_i^{t+1},\hat x^t_i) + \mathcal D^t_{\hat x_i^t}(x_i,x_i^{t+1}).
  \end{equation}
  Hence, from \eqref{iBMD-ie1} and \eqref{iBMD-ie2}, for all $x_i\in\mcalX_i$, we have 
  \begin{equation}
  \label{upperbound}
          f_i^t(x_i^{t+1})  \leq \varphi(x_i) + \mathcal D^t_{\hat x_i^t}(x_i,\hat x_i^t) -  \mathcal D^t_{\hat x_i^t}(x_i,x_i^{t+1}).
  \end{equation}
On the other hand, since $f_i^t(\cdot)$ is convex, we have  $\varphi(x_i)\leq f_i^t(x_i)$. Therefore, we obtain the following inequality for all $x_i\in\mcalX_i$
  \begin{equation}
  \label{iBMD-ie3}
          f_i^t(x_i^{t+1})  +  \mathcal D^t_{\hat x_i^t}(x_i,x_i^{t+1})\leq f_i^t(x_i) + \mathcal D^t_{\hat x_i^t}(x_i,\hat x_i^t).
  \end{equation}
Taking $x_i=x_i^t$ in \eqref{iBMD-ie3}, using  the assumption in~\eqref{Ci-upper}, and summing up the inequalities from $i=1$ to $s$, we obtain  
\begin{equation*}
    f(x^{t+1}) + \sum_{i=1}^s  \mathcal D^t_{\hat x_i^t}(x^t_i,x_i^{t+1})\leq f(x^t) + \sum_{i=1}^s  C_i (\alpha_i^t)^2 \|\mcalP_i(x_i^t-x_i^{t-1})\|^2
\end{equation*}
which further implies the following inequality for all $T\geq 1$
\begin{equation}
\label{iBMD-ie4}
  f(x^{T+1}) +   \sum_{t=1}^T \sum_{i=1}^s  \mathcal D^t_{\hat x_i^t}(x^t_i,x_i^{t+1})\leq f(x^1) +  \sum_{t=1}^T\sum_{i=1}^s  C_i (\alpha_i^t)^2 \|\mcalP_i(x_i^t-x_i^{t-1})\|^2. 
\end{equation}
 From \eqref{iBMD-ie4} and the condition in \eqref{converge_condition-iBMD} we have \eqref{quadratic-series} and $\{f(x^t)\}_{t\geq 0}$ is bounded. Hence $\{x^t\}_{t\geq 0}$ is also bounded if $f$ is assumed to have bounded level sets.

Suppose $x^*$ is a limit point of $\{x^k\}$, that is, there exist a subsequence $\{x^{t_k}\}$ converging to $x^*$. From \eqref{quadratic-series}, we have  $\mathcal D^t_{\hat x_i^t}(x^t_i,x_i^{t+1})\to 0$. Hence, $\{x^{t_k+1}\}$ also converges to $x^*$.   
On the other hand, as $\alpha_i^t \|\mcalP_i(x_i^t-x_i^{t-1})\| \to 0$, we have $\hat x_i^t \to x_i^*$. From the update in \eqref{eq:BMDstep}, 
$$G^{t_k}_i(x^{t_k+1}_i,\hat x_i^{t_k}) \leq G^{t_k}_i(x_i,\hat x_i^{t_k}), \quad \forall x_i \in \mcalX_i
$$
Taking $t_k\to\infty$, from Condition (C1), we have 
$$\bar G_i(x_i^*,x_i^*) \leq \bar G_i(x_i,x_i^*), \quad \forall x_i \in \mcalX_i, $$
where $\bar G_i(\cdot,\cdot)$ is a majorizer of $x_i\mapsto f_i^*(x_i)=(x_1^*,\ldots,x_{i-1}^*,x_i,x_{i+1}^*,\ldots,x_s^*)$. Hence, 
$$ 0\in \partial (\bar G_i(x_i^*,x_i^*) + I_{\mcalX_i} (x_i^*)), \quad \text{for}\, i=1,\ldots,s.$$
Finally, using Proposition  \cite[Proposition 2.1]{Attouch2010} 
and noting that $\nabla_1 \bar G_i(x_i^*,x_i^*)= \nabla_i f(x^*)$, this implies that $x^*$ is a critical point of \eqref{eq:compositev2}. 
\end{proof}
}

\paragraph{Complexity, scalability, and practical implementation aspects} BMMe serves as an accelerated version of the block majorization-minimization method (BMM). 
In essence, while BMM updates each block $x_i$ by  $x^{t+1}_i = \argmin_{x_i\in \mathcal X_i} G^t_i(x_i, x_i^t)$, BMMe achieves the same by replacing $x_i^t$ by an extrapolation point $\hat x_i^t$, which is computed with a marginal additional cost, namely $O(n)$ operations where 
$n$ is the number of variables.  
Consequently, BMMe inherits crucial properties regarding complexity, scalability, and practical implementation from BMM. BMM updates one block of variables at a time while keeping others fixed, thus scaling effectively with data size in terms of the number of block variables. However, the complexity of each block update grows with the dimension of the block variable. The efficiency of BMM heavily relies on selecting suitable majorizers, ensuring closed-form solutions for block updates, thus circumventing the need for outer solvers in large-scale problems (this may help avoid substantial computational resources). The choice of appropriate majorizers is pivotal and application-specific; for instance, in applications utilizing the $\beta$-NMF model, Jensen majorizers are commonly and effectively employed; see Section~\ref{sec:betanmf} below for the details. 
Moreover, it is worth noting that a properly-designed majorizer allows for the computation of its closed-form minimizer in an element-wise manner. This characteristic 
is particularly beneficial for handling large-scale problems, as it can be efficiently executed on a parallel computation platform. 

\revision{ 
\paragraph{Iteration complexity} 
Iteration complexity of BMM-type methods is a challenging topic, especially for nonconvex problems. Considering the use of general majorizers together with inertial/extrapolated parameters in each block update, the most related work to our paper is~\cite{HPGTITAN}. As explained in \cite[Remark~9]{HPGTITAN}, as long as a global convergence (that is, the whole generated sequence converges to a critical point) is guaranteed, a convergence rate for the generated sequence can be derived by using the same technique as in the proof of \cite[Theorem~2]{attouch2009convergence}. 
In fact, the technique of \cite{attouch2009convergence} has been commonly used to establish the convergence rate in other block coordinate methods, for which specific majorizers are used in each block update (for example, \cite{Xu2013} uses Lipschitz gradient majorizers).  
However, it is challenging to extend the result to BMMe. Along with the Kurdyka-Lojasiewicz assumption, the BMM-type methods with extrapolation, such as \cite{HPGTITAN, Xu2013}, need to establish the NSDP, as discussed in Section~\ref{sec:prevworks}. 
And as such, extending the usual convergence rate result to BMMe is an open question.  A recent work~\cite{lyu2020block} establishes iteration complexity of a BMM method for solving constrained nonconvex nonsmooth problems; however the majorizers are required to have Lipschitz gradient, which is not satisfied by many Bregman majorizers. 
}

\section{Application of BMMe to $\beta$-NMF} 
\label{sec:betanmf} 
\revise{Before presenting the application of BMMe to the standard $\beta$-NMF problem with $\beta\in [1,2]$ (Section \ref{sec:beta-nmf}), and a constrained and regularized KL-NMF problem (Section~\ref{sec:KLnmf}),}
we briefly discuss the majorizers for the $\beta$ divergences.

\subsection{Majorizer of the $\beta$ divergence, $\beta\in [1,2]$} \label{sec:majorNMFbeta}

\revise{Recall that the function defined in \eqref{betaNMF-majorizer}} 
is a majorizer for $h\mapsto D_{\beta}(v,Wh)$, where $\tilde v=W\tilde{h}$, see Proposition~\ref{prop:majorizer}. 
The function $g(\cdot,\cdot)$ is twice continuously differentiable over $\{(h,\tilde h): h\geq\varepsilon, \tilde h\geq \varepsilon\}$, and 
\begin{equation}
\label{hessian-hk}
\nabla^2_{h_k} g (h,\tilde h)=\sum_{i=1}^m\frac{W_{ik} \tilde {v}_{i}}{\tilde {h}_{k} }  d''_\beta\left(v_i, \tilde  v_i \frac{h_k}{\tilde  h_k} \right),
\end{equation}
where $d''_\beta(x,y)$ denotes the second derivative with respect to $y$ of $(x,y)\mapsto d_\beta(x,y)$. 

\revise{On the other hand,} as already noted in Section~\ref{sec:prel-MU}, 
 \[ 
 D_{\beta}(X,WH)=\sum_{j=1}^n D_{\beta}(X_{:j},WH_{:j})=\sum_{i=1}^m D_{\beta}(X_{i:}^\top,H^\top W_{i:}^\top).
 \] 
Hence a majorizer of $H\mapsto D_{\beta}(X,W H)$ while fixing $W$ is given by 
\begin{equation}
    \label{eq:H_majorizer-beta} G^{(W)}_2(H,\tilde{H})=\sum_{j=1}^n g^{(W)}_j(H_{:j},\tilde{H}_{:j}),
\end{equation} 
where $g^{(W)}_{j}(H_{:j},\tildeH_{:j})$ is the majorizer of $H_{:j}\mapsto D_{\beta}(X_{:j},WH_{:j})$, which is defined as in \eqref{betaNMF-majorizer} with $v=X_{:j}$.  Similarly, the following function is a majorizer of $W\mapsto D_{\beta}(X,W H)$ while fixing $H$
\begin{equation} 
\label{eq:W_majorizer-beta}
G^{(H)}_1(W,\tilde{W})=\sum_{i=1}^m \mathbf g^{(H)}_i(W_{i:}^\top,\tilde{W}_{i:}^\top) , 
\end{equation}
where $\mathbf g^{(H)}_i(W_{i:}^\top,\tilde{W}_{i:}^\top)$ is the majorizer of $W^\top_{i:}\mapsto D_{\beta}(X_{i:}^\top,H^\top W^\top_{i:})$ defined as in~\eqref{betaNMF-majorizer} with $v$ being replaced by $X_{i:}^\top$ and $W$ being replaced by $H^\top$.

\ngir{Note that there exist other majorizers for $\beta$-NMF, for example majorizers for both variables simultaneously~\cite{marmin2023joint}, 
for $\ell_1$-regularized $\beta$-NMF with sum-to-one constraints~\cite{marmin2023majorization}, 
and 
quadratic majorizers for the KL divergence~\cite{pham2023fast}. 
} 
\ngiii{
 \begin{remark}[Choice of majorizer] 
     Incorporating a regularization term $\lambda |x_i - x_i^t|^2$ in the Jensen surrogate to have a strongly convexity majorizer 
     would fulfill the conditions outlined in TITAN~\cite{HPGTITAN}. However, this is not \ngi{recommended}, as it would result in a regularized Jensen surrogate that lacks a closed-form solution for the subproblem, necessitating an outer solver and requiring adaptation of the convergence analysis of TITAN to accommodate inexact solutions (the current analysis of TITAN does not support inexact solutions). In contrast, the extrapolation strategy employed by BMMe will preserve the closed-form update for $\beta$-NMF in its iterative step by embedding the extrapolation point directly into the majorizer. 
 \end{remark}
} 
 
\subsection{MU with extrapolation for $\beta$-NMF, $\beta\in [1,2]$} 
\label{sec:beta-nmf} 

\revise{We consider the standard $\beta$-NMF problem in \eqref{betaNMF}} with $\beta\in[1,2]$. 
Applying Algorithm~\ref{alg:BMMe} to solve Problem~\eqref{betaNMF}, we get MUe, a multiplicative update method with extrapolation described in Algorithm~\ref{alg:MUe-beta}. Convergence property of MUe is given in Theorem~\ref{thrm:MUe-beta}; see Appendix~\ref{app:proof41} for the proof. 

\begin{algorithm}[ht!]
\caption{MUe for solving $\beta$-NMF~\eqref{betaNMF}} 
\label{alg:MUe-beta}
\begin{algorithmic}[1]
\STATE Choose initial points $(W^{-1}, W^0, H^{-1}, H^0) \geq \varepsilon \ngir{ > 0}$.
\FOR{$t=\ngi{0},\ldots$}

\STATE Compute extrapolation points: 
$$\begin{array}{ll}
\hatW^t=W^t + \alpha_W^t[W^t-W^{t-1}]_+, \\
\hatH^t=H^t + \alpha_H^t[H^t-H^{t-1}]_+, 
\end{array} 
$$
where $\alpha_W^t$ and $\alpha_H^t$ satisfy \eqref{converge_condition-beta}.

\STATE Update the two blocks of variables: 
\begin{align*}
W^{t+1} & = \underset{W\geq\varepsilon}{\argmin} G^t_1(W,\hatW^t) = 
\text{MU}\big(  X^\top , (H^t)^\top, (\hat W^t)^\top \big)^\top \text{ [see \eqref{eq:standardMU}], }  
\\
H^{t+1} &  = \underset{H\geq\varepsilon}{\argmin} G^t_2(H,\hatH^t)
\; = \text{MU}\big(X,W^{t+1},\hat H^t\big), 
\end{align*} 
where \revise{$G_1^t=G_1^{(H^{t})}$ and $G^t_2=G_2^{(W^{t+1})}$ be the majorizers defined in \eqref{eq:W_majorizer-beta} with $H=H^{t}$ and \eqref{eq:H_majorizer-beta} with $W=W^{t+1}$, respectively. 
} 
\ENDFOR
\end{algorithmic}
\end{algorithm} 

\begin{theorem}
\label{thrm:MUe-beta}
    Suppose the extrapolation parameters \ngi{in Algorithm~\ref{alg:MUe-beta}} are chosen such that they are bounded and  
    \begin{equation} \label{converge_condition-beta}
        \sum_{t=1}^{\infty} (\alpha^t_{H})^{2}\big\|[H^{t}-H^{t-1}]_{+} \big\|^{2}<+\infty, 
       \; \text{ and }  \; 
       \sum_{t=1}^{\infty}(\alpha^t_{W})^{2} \big\|[W^{t}-W^{t-1}]_{+}\big\|^{2}<+\infty. 
        \end{equation}
        Then MUe (Algorithm~\ref{alg:MUe-beta}) 
        generates a bounded sequence and any of its limit point, $(W^*,H^*)$, is a KKT point of Problem~\eqref{betaNMF}, that is,  
 \begin{equation*}
    W^*\geq\varepsilon, \; 
    \nabla_W D_\beta (X,W^*H^*)\geq 0, \;  
    \langle \nabla_W D_\beta (X,W^*H^*), W^* - \varepsilon e e^\top\rangle = 0, 
 \end{equation*}
 \ngir{and similarly for $H^*$.}  
\end{theorem}


\subsection{MUe for constrained and regularized KL-NMF} \label{sec:KLnmf}

The KL divergence is especially relevant when the statistical characteristics of the observed data samples conform to a Poisson distribution, turning KL-NMF into a meaningful choice for count data sets such as \ngi{images~\cite{richardson1972bayesian}, documents~\cite{Lee99},  and single-cell sequencing~\cite{carbonetto2023gom}}. 
In numerous scenarios, there are specific \ngi{additional constraints and regularizers to add} to KL-NMF. 
For example, the minimum-volume (min-vol) KL-NMF, which incorporates a regularizer encouraging  the columns of matrix $W$ to have a small volume, along with a normalization constraints (such as $H^\top e=e$ or $W^\top e = e$), enhances identifiability/uniqueness~\cite{lin2015identifiability,fu2015blind}, 
a crucial aspect in various applications such as hyperspectral imaging~\cite{miao2007endmember} and audio source separation~\cite{leplat2020blind}. 
In this section, we consider the following general regularized KL-NMF problem 
 \begin{equation} 
 \label{u_KLNMF}
\underset{W\in\bar{\Omega}_W, H\in\bar\Omega_H} {\min}\Big\{f(W,H)\, := \, D_{KL}(X,WH) + \lambda_1\phi_1(W) + \lambda_2\phi_2(H)\Big\},
\end{equation}
where $
    \bar{\Omega}_W:=\big\{W:W\geq\varepsilon, W\in\Omega_W\big\}$ and $
    \bar\Omega_H:=\big\{H:H\geq\varepsilon, H\in\Omega_H\big\}.
$ 
We assume that  there exist continuous functions $L_{\phi_1}(\tilde W)\geq 0$ and $L_{\phi_2}(\tilde H)\geq 0$ such that for all $W, \tildeW\in\{W:W\geq 0, W\in\Omega_W\}$ and $H,\tildeH\in\{H: H\geq 0, H\in \Omega_H\}$, we have
\begin{equation}
\label{quadratic_seperable}
    \begin{array}{ll}
\phi_1(W) &\leq \; \barphi_1(W,\tilde W):= \phi_1(\tilde W) + \langle \nabla \phi_1(\tilde W), W-\tilde W\rangle + \frac{L_{\phi_1}(\tilde W)}{2}\|W-\tilde W\|^2, \\ \vspace{0.2cm} 
\phi_2(H) &\leq \; \barphi_2(H,\tilde H):= \phi_2(\tilde H) + \langle \nabla \phi_2(\tilde H), H-\tilde H\rangle + \frac{L_{\phi_2}(\tilde H)}{2}\|H-\tilde H\|^2.
\end{array}
\end{equation}
Furthermore, $L_{\phi_1}(\tilde W)$ and $L_{\phi_2}(\tilde H)$ in \eqref{quadratic_seperable} are upper bounded by $\bar L_{\phi_1}$ and $\bar L_{\phi_2}$, respectively. 
\ngi{We focus on the min-vol regularizer, $\phi_1(W) = \det(W^\top W + \delta I)$, and \mbox{$\Omega_W = \{W \ | \ W^\top e =e \}$} \cite{leplat2020blind}. In that case,  $\phi_1(W)$  satisfies this condition, as proved in Lemma~\ref{lemma:minvol}; see Appendix~\ref{proof-minvol} for the proof.} 
\begin{lemma}
    \label{lemma:minvol}
   The function $\phi_1(W)=\log\det(W^\top W+\delta I)$ with $\delta > 0$ satisfies the condition in \eqref{quadratic_seperable} with $L_{\phi_1}(\tildeW)=2\|(\tilde{W}^\top \tilde{W}+\delta I)^{-1} \|_2$, which is upper bounded by \revision{$2/\delta$}.
\end{lemma}
We will use the following majorizer for $H\mapsto f(W,H)$ while fixing $W$: 
\begin{equation}
    \label{eq:H_majorizer} 
    G^{(W)}_2(H,\tilde{H}) =\sum_{j=1}^n g^{(W)}_j(H_{:j},\tilde{H}_{:j}) 
      + \lambda_1  \phi_1(W) + \lambda_2 \bar{\phi}_2(H,\tilde H),
\end{equation}
where $\bar{\phi}_2(H,\tilde H)$ is defined in~\eqref{quadratic_seperable} and $g^{(W)}_{j}$ is \revise{defined as in \eqref{eq:H_majorizer-beta}.} 
Similarly, we use the following  majorizer for $W\mapsto f(W,H)$ while fixing $H$: 
\begin{equation} 
\label{eq:W_majorizer} 
G^{(H)}_1(W,\tilde{W}) =\sum_{i=1}^m \mathbf g^{(H)}_i(W_{i:}^\top,\tilde{W}_{i:}^\top) + \lambda_1 \bar{\phi}_1(W,\tilde W) + \lambda_2 \phi_2(H), 
\end{equation}
where $\bar{\phi}_1(W,\tilde W)$ is defined in~\eqref{quadratic_seperable}, and $\mathbf g^{(H)}_i$ is \revise{defined as in \eqref{eq:W_majorizer-beta}.} 
Applying Algorithm~\ref{alg:BMMe} to solve Problem~\eqref{u_KLNMF}, we get Algorithm~\ref{alg:MUe-KL}, a BMMe algorithm for regularized and constrained KL-NMF \revise{see its detailed description in Appendix~\ref{sec:MUe-KL}. 
\ngir{It works exactly as} Algorithm~\ref{alg:MUe-beta} but the majorizers $G_1^t=G_1^{(H^{t})}$ and $G^t_2=G_2^{(W^{t+1})}$ are defined in \eqref{eq:W_majorizer} with $H=H^{t}$ and \eqref{eq:H_majorizer} with $W=W^{t+1}$, respectively.  }  Convergence property of Algorithm~\ref{alg:MUe-KL} is given in Theorem \ref{thrm:MUe-kl}; see Appendix~\ref{app:th43} for the proof. 
\begin{theorem}
\label{thrm:MUe-kl}
    Suppose the extrapolation parameters \revise{$\alpha_W^ t$ and $\alpha_H^t$} \ngi{satisfy} \eqref{converge_condition-beta}. 
        Then \revise{BMMe applied to 
 Problem~\eqref{u_KLNMF}} (Algorithm~\ref{alg:MUe-KL}) generates a bounded sequence and any of its limit point is a coordinate-wise minimizer of Problem~\eqref{u_KLNMF}. 
\end{theorem}

\ngi{
\revise{BMMe for solving Problem~\eqref{u_KLNMF} (Algorithm~\ref{alg:MUe-KL})} is not necessarily straightforward to implement, because \revise{its updates 
might not have closed forms}. \revise{In the following, we  derive such updates in the special case of min-vol KL-NMF~\cite{leplat2020blind}: 
}}
\begin{align}
\min_{W\geq\varepsilon,H\geq\varepsilon} & D_{KL} (X,WH) + \lambda_1\log\det(W^\top W+\delta I) 
\nonumber \\
& \text{ such that } 
\quad e^\top W_{:j}=1, j=1,\ldots,r. \label{min-vol-klnmf}  
\end{align} 
The update of $H$ is as in~\eqref{eq:standardMU} taking $\beta = 1$. The following lemma provides the update of $W$; see Appendix~\ref{proof-updatew} for the proof. 
\begin{lemma}
\label{lemma:updatew}
    For notation succinctness, let $\hat W= W^t + \alpha_W^t[W^t-W^{t-1}]_+$ and $H=H^{t+1}$. 
    BMMe for solving \eqref{min-vol-klnmf} (Algorithm~\ref{alg:MUe-KL}) updates $W$ as follows: 
   \begin{equation}\label{iMU_Wupdate} 
   W \leftarrow 
   \max\Big( \varepsilon,\frac12 \big(-B_2 + \Big [[B_2]^{.2} + 4 \lambda_1  L_{\phi_1}(\hat W) B_1 \Big]^{.1/2} \big)  \Big),
   \end{equation}
where 
$$
\begin{array}{ll}
 & L_{\phi_1}(\hat W)=2\|(\hat{W}^\top \hat{W}+\delta I)^{-1} \|_2, \; 
 B_1=\frac{[X]}{[\hat W H]}H^\top \circ \hat{W},\\
&
 B_2=e e^\top H^\top + \lambda_1 \big(A - L_{\phi_1}(\hat W) \hat W  + e \mu^\top\big), \\
 &A=2\hat W(\hat{W}^\top \hat{W}+\delta I)^{-1}, \quad\mu=(\mu_1,\ldots,\mu_r)^\top,
\end{array}$$
and $\mu_k$, for $k=1,\ldots,r$, is the unique solution of $\sum_{j=1}^n  W_{jk}(\mu_k)=1$. We can determine $\mu_k$ by using bisection method over  $\mu_k\in [\underline{\mu}_k,\overline{\mu}_k]$, where 
\begin{equation}
\label{mujk}
\begin{array}{ll}
\ngir{\underline{\mu}_k} & \ngir{ = 
\underset{j=1,\ldots,m}{\min} \tilde\mu_{jk}, \quad 
\overline{\mu}_k = 
\underset{j=1,\ldots,m}{\max} \tilde\mu_{jk}, } 
\\
\tilde\mu_{jk}&=\frac{1}{\lambda_1}(4 \lambda_1  L_{\phi_1}(\hat W)  b_1 m -1/m-\sum_{i=1}^n (H^\top)_{ik}) +L_{\phi_1}(\hat W) \hat W_{jk}-A_{jk},
\end{array}
\end{equation}
with $b_1=\sum_{i=1}^n\frac{(H^\top)_{ik} X_{ji}}{\tilde{v}_{i}}  \hat{W}_{jk}$ and $\tilde v=H^\top \hat W_{j:}^\top $. 
\end{lemma}

\section{Numerical experiments} 
\label{sec:experiment}

In this section, we show the empirical acceleration effect of BMMe. We use the Nesterov extrapolation parameters~\eqref{eq:exrapolNes}. 
All experiments have been performed on a laptop computer with Intel Core i7-11800H @ 2.30GHz and 16GB memory with MATLAB R2021b. The code is available from \url{https://github.com/vleplat/BMMe}. 

 
\subsection{$\beta$-NMF for hyperspectral imaging} 

\begin{figure}[ht!]
\begin{center}
\includegraphics[width=0.6\textwidth]{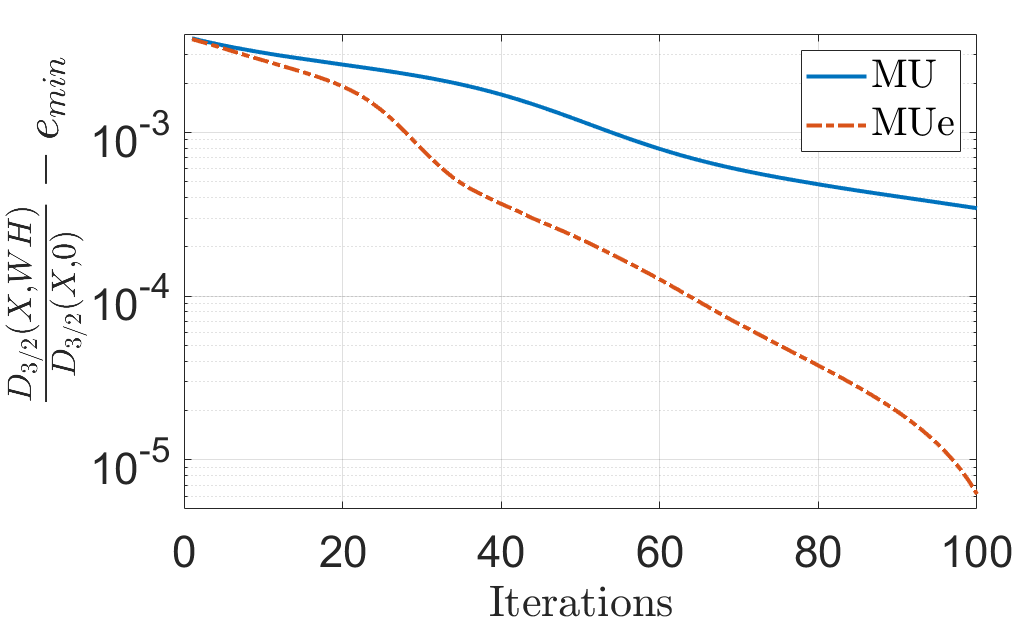} 
\caption{Median relative objective function of $\beta$-NMF for $\beta = 3/2$ minus the smallest objective  function found among 10 random initializations (denoted $e_{min}$).}
\label{fig:cuprite}
\end{center}
\end{figure}

We consider $\beta$-NMF~\eqref{betaNMF} with $\beta = 3/2$ which is among the best NMF models for hyperspectral unmixing~\cite{fevotte2014nonlinear}.  
For this problem, the MU are the workhorse approach, and we compare it to MUe: Figure~\ref{fig:cuprite} provides the median evolution of objective function values for the Cuprite data set ($m=188$, $n = 47750$, $r=20$); see the Supplementary Material SM1 for more details and experiments on 3 other data sets with similar observations.  
There is a significant acceleration effect: on average, MUe requires only 41 iterations to obtain a smaller objective than MU with 100 iterations.

\subsection{KL-NMF for topic modeling and imaging} 

\begin{figure}[ht!]
\begin{center}
\begin{tabular}{cc}
     \includegraphics[width=0.47\textwidth]{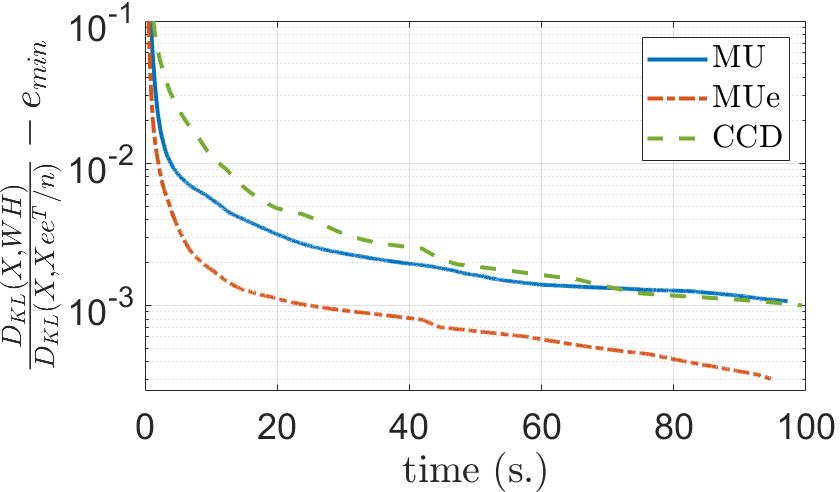}   & 
     \includegraphics[width=0.47\textwidth]{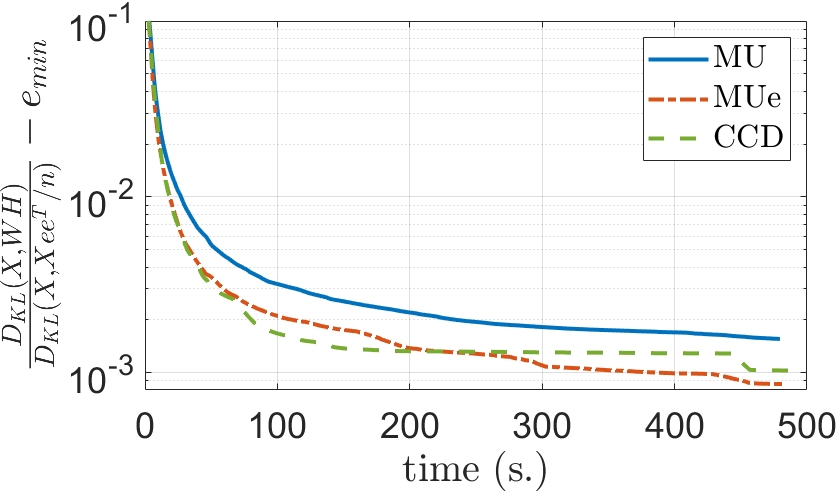}  
\end{tabular} 
\caption{Median relative objective function of KL-NMF minus the smallest objective  function found among 10 runs (denoted $e_{min}$) \revision{w.r.t. CPU time}: (\revision{left}) ORL, (\revision{right}) hitech, with $r=10$ in both cases. \revision{Note that MU, MUe and CCD respectively perform on average 7579, 6949 and 616 iterations for ORL, and 885, 886 and 343 iterations for hitech in the considered time intervals.}\label{fig:KLNMFexp}} 
\end{center}
\end{figure}

We now consider KL-NMF 
which is the workhorse NMF model for topic modeling and also widely used in imaging; see Section~\ref{sec:KLnmf}. 
We compare MUe with MU and the 
cyclic coordinate descent (CCD) method of  \cite{Hsieh2011}. As reported in~\cite{HienNicolasKLNMF}, MU and CCD are the state of the art for KL-NMF (sometimes one performs best, sometimes the other).    
Figure~\ref{fig:KLNMFexp} shows the median evolution of the \revise{relative} objective function 
for two data sets:  a dense image data set (ORL, $m=10304$, $n=400$), and a sparse document data set (hitech, $m=2301$, $n=10080$).  
For ORL, CCD and MU perform similarly while MUe performs the best. 
For hitech, MUe and CCD perform similarly, while they outperform MU. In all cases, MUe provides a significant acceleration effect to MU. 
Similar observations hold for 6 other data sets; see the Supplementary Material SM2. 


\subsection{Min-vol KL-NMF for audio data sets} \label{sec:experminvolKL} 

We address Problem~\eqref{min-vol-klnmf} in the context of blind audio source separation \cite{leplat2020blind}. 
We compare the following algorithms: 
MU with $H$ update from~\eqref{eq:standardMU} and $W$ update from Lemma~\ref{lemma:updatew}, its extrapolated variant, MUe, 
a recent MM algorithm by~\cite{Leplat2021}, denoted MM, 
and MMe incorporating the BMMe extrapolation step in MM.  

\begin{figure}[ht!]
\begin{center}
\begin{tabular}{c}
\includegraphics[width=0.47\textwidth]{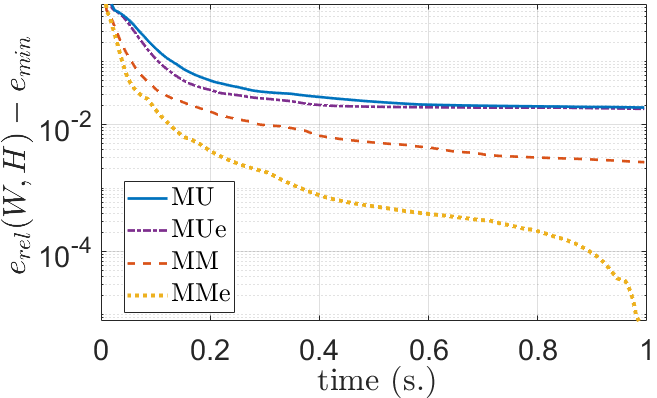} 
\end{tabular}
\caption{Evolution of the median relative errors (see~\eqref{eq:rel_error_minvolklnmf}) minus the smallest relative error found among the 20 random  initializations, of min-vol KL-NMF ~\eqref{min-vol-klnmf} of the four algorithms. \revision{Note that MU, MUe, MM and MMe respectively perform on average 94, 92, 304 and 278 iterations within one second.}}  
\label{fig:prelude}
\end{center}
\end{figure}

Figure~\ref{fig:prelude} displays the median relative objective function values, 
\begin{equation}\label{eq:rel_error_minvolklnmf}
    e_{rel}(W,H) = \frac{D_{KL}(X,WH) + \lambda \log\det(W W+\delta I)}{D_{KL}(X, (X e/n) e^\top)}, 
\end{equation} 
minus the smallest relative objective found,  
for the prelude from J.S.-Bach ($m=129$, $n = 2292$, $r=16$); see the Supplementary Material SM3 for two other data sets, and more details. 
MUe exhibits accelerated convergence compared to MU. MM ranks second, while its new variant, MMe,  integrating the proposed extrapolation, consistently achieves the best performance. 
This better performance stems from the nature of the majorizer employed for the logdet term which offers a more accurate approximation. 
\revise{It is worth noting that} the majorizer used by MMe \revise{neither} satisfies Condition~(c) of Definition~\ref{def:majorizer} \revise{nor Assumption 1 (A3) of  \cite{Razaviyayn2013}}, and hence Theorem~\ref{thrm:BMMe} does not apply to MMe. 



\section{Conclusion and further work}
\label{sec:conclusion}

In this paper, we considered multi-convex optimization~\eqref{eq:compositev2}. 
We proposed a new simple \revise{yet effective} acceleration mechanism for the block majorization-minimization method, \revise{incorporating extrapolation} (BMMe). 
We \revise{established} 
subsequential convergence of BMMe, and \revise{leveraged} 
it to accelerate multiplicative updates for various NMF problems. 
\revise{Through numerous numerical experiments conducted on diverse datasets, namely documents, images, and audio datasets, we showcased the remarkable acceleration impact achieved by BMMe}.
Further work include the use of BMMe for other NMF models and algorithms~\cite{marmin2023joint, marmin2023majorization} and other applications, \revise{such as} nonnegative tensor decompositions, 
and new theoretical developments, such as relaxing Condition (c) in Definition~\ref{def:majorizer} (definition of a majorizer) or the condition on directional derivatives \cite[Assumption 1 (A3)]{Razaviyayn2013}, \revision{extending to the case when the subproblems in each block of variables are not convex\footnote{We need the convexity assumption for the proof of Theorem~\ref{thrm:BMMe} and Proposition~\ref{prop:threepoint}. Without convexity, we cannot establish the inequality~\eqref{iBMD-ie4}, which is key to prove Theorem~\ref{thrm:BMMe}.}, or studying iteration complexity of BMMe (see the paragraph at the end of Section~\ref{sec:BMMe}).} 



\revision{
\section*{Acknowledgments}  We are grateful to the anonymous reviewers who
carefully read the manuscript, their feedback helped us improve our paper.  
}

\appendix

\section{Technical proofs} \label{app:techproofs}

\subsection{Proof of Proposition~\ref{prop:threepoint}} \label{sec:propthreepoint}

To prove our convergence result for BMMe in Theorem~\ref{thrm:BMMe}, we need the following useful proposition which is an extension of Property 1 of \cite{Tseng2008}. 

\begin{proposition}\label{prop:threepoint}
Let $ z^+=	\arg\min_{u \in \mathcal Y} \varphi(u)+\mathcal B_\xi(u,z)$, where $\varphi$ is a proper convex function, $\mathcal Y$ is a closed convex set, and $\mathcal B_\xi(u,z)=\xi(u)-\xi(z)-\langle \nabla \xi(z),u-z\rangle$, where $\xi$ is a convex differentiable function in $u$ while fixing $z$ (note that $\xi$ may also depend on $z$, and we should use $\xi^{(z)}$ for $\xi$ but we omit the upperscript for notation succinctness). Then for all $u\in \mathcal Y$ we have 
\begin{align*}
\varphi (u) + \mathcal B_\xi(u,z) \geq \varphi(z^+) + \mathcal B_{\xi}(z^+,z) + \mathcal B_{\xi}(u,z^+).
\end{align*}
\end{proposition}
\begin{proof}
Optimality condition gives us 
\begin{align*}
\langle \varphi'(z^+)+\nabla_1 \mcalB_\xi(z^+,z),u-z^+\rangle \geq 0, \forall u\in \mathcal Y,
\end{align*}
where $\varphi'(z^+)$ is a subgradient of $\varphi$ at $z^+$. Furthermore, as $\varphi$ is convex, we have 
\begin{align*}
\varphi(u)\geq \varphi(z^+) + \langle \varphi'(z^+),u-z^+\rangle.
\end{align*}
Hence, for all $u\in \mathcal Y$, 
\begin{align*}
\varphi (u) + \mathcal B_\xi(u,z) 
& \geq  \varphi(z^+)-\langle\nabla_u B_\xi(z^+,z),u-z^+\rangle +  \mathcal B_\xi(u,z) \\
&=\varphi(z^+)-\langle \nabla\xi(z^+)-\nabla\xi(z),u-z^+\rangle +\xi(u)-\xi(z)-\langle \nabla \xi(z),u-z\rangle\\
&=\varphi(z^+)+\mathcal B_{\xi}(z^+,z) + \mathcal B_{\xi}(u,z^+). 
\end{align*}
\end{proof}

\subsection{Proof of Theorem \ref{thrm:MUe-beta}} \label{app:proof41} 

    Let us first prove that the generated sequence of Algorithm~\ref{alg:MUe-beta} is bounded.
    For simplicity, we denote $W=W^{t+1}$ and $\hatH=\hatH^t$ in the following. We have
 \begin{align*}
     H^{t+1}_{kj}&=\hatH_{kj}\frac{\sum_{i=1}^m W_{ik} \frac{X_{ij}}{([W \hatH]_{ij})^{2-\beta}} }{\sum_{i=1}^m W_{ik} ([W \hatH]_{ij})^{\beta-1} } =\sum_{i=1}^m \frac{\hatH_{kj} W_{ik} \frac{X_{ij}}{([W \hatH]_{ij})^{2-\beta}} }{\sum_{i=1}^m W_{ik} ([W \hatH]_{ij})^{\beta-1}}\\
   &  \leq \sum_{i=1}^m \frac{\hatH_{kj} W_{ik} \frac{X_{ij}}{([W \hatH]_{ij})^{2-\beta}} }{W_{ik} ([W \hatH]_{ij})^{\beta-1}}=\sum_{i=1}^m \hatH_{kj}  \frac{X_{ij}}{[W \hatH]_{ij}}=\sum_{i=1}^m \hatH_{kj}  \frac{X_{ij}}{\sum_{l=1}^{r} W_{il} \hatH_{lj}}\\
   &\leq \sum_{i=1}^m  \hatH_{kj}  \frac{X_{ij}}{W_{ik} \hatH_{kj}} \leq  \sum_{i=1}^m \frac{X_{ij}}{\varepsilon}.
 \end{align*}   
 Hence, $\{H^t\}_{t\geq 0}$ is bounded. 
 Similarly we can prove that $\{W^t\}_{t\geq 0}$ is bounded. 
 
 Now we verify the conditions of Theorem~\ref{thrm:BMMe}. Note that $W\mapsto G_1^{(H)}(W,\tildeW)$ and $H\mapsto G_2^{(W)}(H,\tildeH)$ are convex.

 \paragraph{Condition (C1) of Theorem~\ref{thrm:BMMe}} We see that $(W,H,\tildeH)\mapsto G_2^{(W)}(H,\tildeH)$ is continuously differentiable over $\{(W,H,\tildeH):W\geq\varepsilon, H\geq\varepsilon, \tildeH\geq\varepsilon\}$. Furthermore, suppose $(W^{t+1},H^t)\to (\bar W,\bar H)$ then we have $\bar W\geq\varepsilon$ and $\bar H\geq\varepsilon$ as $W^{t+1}\geq\varepsilon$ and $H^t\geq\varepsilon$. Hence, it is not difficult to verify that $G_2^t$ satisfies Condition (C1) of Theorem~\ref{thrm:BMMe}, and similarly for $G_1^t$.

 \paragraph{Condition (C2) of Theorem~\ref{thrm:BMMe}} Considering Condition (C2) of Theorem~\ref{thrm:BMMe}, if we fix $\bar x$ and $\tildex_i\in \mcalY_i$, where $\mcalY_i$ is a closed convex set, and $G^{(\bar x)}_i(\cdot,\tildex_i)$ is twice continuously differentiable over $\mcalY_i$ and the norm of its Hessian is upper bounded by $2C_i$ over $\mcalY_i$,  then by the descent lemma~\cite{NesterovLecture2018}, we have
 \begin{align*}
 G^{(\bar x)}_i(x_i,\tildex_i)&\leq \bar f_i(\tildex_i)+ \langle \nabla \bar f_i(\tildex_i),x_i - \tildex_i\rangle+ C_i\|x_i-\tildex_i\|^2, \forall x_i\in\mcalY_i. 
\end{align*}
 This implies that 
 \begin{align*}
 &\mcalD_{\bar x,\tildex_i}(x_i,\tildex_i)=G^{(\bar x)}_i(x_i,\tildex_i)- \left(\bar f_i(\tildex_i)+ \langle \nabla \bar f_i(\tildex_i),x_i - \tildex_i\rangle\right)\leq C_i  \|x_i - \tildex_i\|^2, \forall x_i\in\mcalY_i, 
\end{align*} 
 and hence that the Condition (C2) is satisfied.

 Now consider Algorithm~\ref{alg:MUe-beta}.  Note that $W^{t}\geq \varepsilon$, $H^{t}\geq \varepsilon$, $\hat H^t=H^t + \alpha_H^t[H^t-H^{t-1}]_+\geq\varepsilon$, $\hat W^t=W^t + \alpha_W^t[W^t-W^{t-1}]_+\geq\varepsilon$, and we have proved that $\{(W^t,H^t)\}_{t\geq 0}$ generated by Algorithm~\ref{alg:MUe-beta} is bounded. This implies that $\{\hat W^t\}_{t\geq0}$  and $\{\hat H^t\}_{t\geq0}$ are also bounded.
   We verify (C2) for block $H$ and it is similar for block $W$; recall that $G_2^{(W)}$ is defined in~\eqref{eq:H_majorizer-beta}. 
  Consider the compact set $\mathcal C=\{(H,\tilde H): H\geq \varepsilon, \tilde H\geq\varepsilon, \|(H,\tilde H)\|\leq C_H\}$, where $C_H$ is a positive constant such that $\mathcal C$ contains $(H^t,\hatH^t)$. Since   $G^{(W)}_2$, with $W\geq\varepsilon$, is twice continuously differentiable over the compact set $\mathcal C$, the Hessian $\nabla^2_H G^{t}_2(\cdot,\hat H^t) $ is bounded by a constant that is independent of $W^t$ and $\hat H^t$. As discussed above, \ngi{this implies that the}  Condition (C2) of Theorem~\ref{thrm:BMMe} is satisfied. 
 

   \paragraph{Condition (C4) of Theorem~\ref{thrm:BMMe}}
 Finally,  from \eqref{hessian-hk}, we see that $ \nabla^2_{h_k} g_j^{(W)} (h,\hat h)$ is lower bounded by a positive constant when $h\geq\varepsilon$, $\hat h\geq \varepsilon$, $W\geq\varepsilon$, and $h$, $\hat h$, and $W$ are upper bounded. Hence the Condition (C4) of Theorem~\ref{thrm:BMMe} is satisfied. 

 By Theorem~\ref{thrm:BMMe}, any limit point $(W^*,H^*)$ of the generated sequence is a coordinate-wise minimizer of Problem \eqref{betaNMF}. Hence, 
 \begin{equation}
 \label{deriveKKT}
 \begin{split}
    W^*\geq\varepsilon,\quad  \langle \nabla_W D_\beta (X,W^*H^*), W-W^*\rangle \geq 0 \quad \forall W\geq \varepsilon, \\
    H^*\geq\varepsilon,\quad \langle \nabla_H D_\beta (X,W^*H^*), H-H^*\rangle \geq 0 \quad \forall H\geq \varepsilon.
 \end{split}
 \end{equation}
By choosing $H=H^*+\mathbf E_{(i,j)}$ in \eqref{deriveKKT} for each $(i,j)$, where $ \mathbf  E_{(i,j)}$ is a matrix with a single component equal to 1 at position $(i,j)$ and the other being 0, we get $\nabla_H D_\beta (X,W^*H^*)\geq 0$.  Similarly, we have  $\nabla_W D_\beta (X,W^*H^*)\geq 0$.  By choosing $H=\varepsilon e e^\top$  and $H=2H^*-\varepsilon e e^\top$ in \eqref{deriveKKT}, we have  $\frac{\partial D_\beta (X,W^*H^*)}{\partial H_{ij}}(\varepsilon-H^*_{ij})=0$. Similarly, we also have $\frac{\partial D_\beta (X,W^*H^*)}{\partial W_{ij}}(\varepsilon-W^*_{ij})=0$. These coincide with the KKT conditions, and hence conclude the proof.

\subsection{Proof of Lemma~\ref{lemma:minvol}}
\label{proof-minvol}
We have   
\begin{equation*}
    \begin{split}
        \phi_1(W)&\leq  \phi_1(\tilde W) + \big\langle (\tilde{W}^\top \tilde{W}+\delta I)^{-1}, W^\top W-\tildeW^\top \tildeW \big\rangle \\
        &\leq  \phi_1(\tilde W) +  \big\langle 2\tildeW(\tilde{W}^\top \tilde{W}+\delta I)^{-1}, W-\tildeW \big\rangle +\|(\tilde{W}^\top \tilde{W}+\delta I)^{-1} \|_2 \|W-\tilde W\|_2^2\\
        &\leq \phi_1(\tilde W) +  \big\langle\nabla\phi_1(\tildeW), W-\tildeW \big\rangle + \|(\tilde{W}^\top \tilde{W}+\delta I)^{-1} \|_2 \|W-\tilde W\|^2,
    \end{split}
\end{equation*}
where we use the concavity of $\log\det(\cdot)$ for the first inequality, and the property that $W\mapsto  \big\langle (\tilde{W}^\top \tilde{W}+\delta I)^{-1}, W^\top W \big\rangle$ is $2\|(\tilde{W}^\top \tilde{W}+\delta I)^{-1} \|_2$-smooth for the second inequality.

\subsection{Proof of Theorem \ref{thrm:MUe-kl}} \label{app:th43}  We verify the conditions of Theorem~\ref{thrm:BMMe}. It is similar to the case of standard $\beta$-NMF with $\beta\in [1,2]$, we have $W\mapsto G_1^{(H)}(W,\tildeW)$ and $H\mapsto G_2^{(W)}(H,\tildeH)$ are convex 
 and Condition (C1) are satisfied. 

    \paragraph{Condition (C2) of Theorem~\ref{thrm:BMMe}}  At iteration $t$, we verify (C2) for block $H$ (recall that  $G_2^{(W)}$ is defined in \eqref{eq:H_majorizer}), and it is similar for block $W$, by symmetry. 
For notation succinctness, in the following we denote $W=W^{t+1}$ and $\hatH=\hatH^t$. Note that $W^{t+1}\geq \varepsilon$ and $\hatH^t=H^t + \alpha_H^t[H^t-H^{t-1}]_+\geq\varepsilon$.  We observe that $G_2^{(W)}$ is separable with respect to the columns $H_{:j}$, $j=1,\ldots,n$, of $H$. Specifically, 
\begin{equation*}
    G^{(W)}_2(H,\tilde{H})=\sum_{j=1}^n \Big( g^{(W)}_j(H_{:j},\tilde{H}_{:j}) + \lambda_2 \bar{\phi}^j_2(H_{:j},\tilde H) \Big)+ \lambda_1  \phi_1(W) ,
\end{equation*}
where $\bar{\phi}^j_2(H_{:j},\tilde H)=\phi_2(\tilde H) + [\nabla \phi_2(\tilde H)]_{:j}^\top (H_{:j}-\tilde H_{:j})+ \frac{L_{\phi_2}(\tilde H)}{2}\|H_{:j}-\tilde H_{:j}\|^2$. 
Hence, as discussed above in the proof of Theorem~\ref{thrm:MUe-beta}, it is sufficient to prove that the norm of the Hessian 
$\nabla^2_h\big(g_j^{(W)} (h,\hatH_{:j}) + \lambda_2 \bar{\phi}^j_2(h,\hatH) \big),
$
for $j=1,\ldots,n$, is upper bounded over $h\geq\varepsilon$ by a constant that is independent of $W^{t+1}$ and $\hatH^t$. As $L_{\phi_2}(\tilde H)$ is assumed to be upper bounded by $\bar L_{\phi_2}$, it is sufficient to prove that $\nabla^2_h g_j^{(W)} (h,\hatH_{:j})$ is upper bounded over $h\geq\varepsilon$.  

We have 
\begin{align*}
\nabla^2_{h_k} g_j^{(W)} (h,\hat h)&=\sum_{i=1}^m\frac{W_{ik} \hat{v}_{i}}{\hat{h}_{k} }  \frac{v_i (\hat h_k)^2}{(\hat v_i h_k)^2}=\sum_{i=1}^m \frac{W_{ik} \hat h_k}{\hat v_i} \frac{v_i}{(h_k)^2} \stackrel{\rm(a)}{\leq} \sum_{i=1}^m \frac{W_{ik} \hat h_k}{\hat v_i}  \frac{v_i}{ \varepsilon^2}   \stackrel{\rm(b)}{\leq}  \sum_{i=1}^m \frac{v_i}{\varepsilon^2},
\end{align*}
where we used $h_k\geq\varepsilon$ in (a) and $ W_{ik} \hat{h}_{k}\leq\hat v_i=\sum_{k=1}^r W_{ik} \hat{h}_{k}$ in (b).
Hence Condition (C2) is satisfied. Together with \eqref{converge_condition-beta}, \ngi{this implies} that the generated sequence of Algorithm~\ref{alg:MUe-KL} is bounded as the objective of \eqref{u_KLNMF} has bounded level sets.

Finally, as the generated sequence is upper bounded and $W\geq\varepsilon$, $H\geq \varepsilon$, and $\hat H\geq \varepsilon$, we see that $\nabla^2_{h_k} g_j^{(W)} (h,\hat h)$ is lower bounded by a positive constant, which implies that the Condition (C4) is satisfied.

\subsection{Proof of Lemma~\ref{lemma:updatew}}
\label{proof-updatew}

The update of $W$ is given by 
\begin{equation}
\label{eq:min-vol-update}
  \begin{split}
W^{t+1}\leftarrow \arg\min_{W\geq\varepsilon} & G_1(W,\hat W)=\sum_{i=1}^m g^i(W_{i:}^\top,\hat W_{i:}^\top) + \lambda_1 \bar{\phi}_1(W,\hat W)
\\
\mbox{s.t.} &\quad e^\top W_{:k}=1, k=1,\ldots,r. 
\end{split}
\end{equation}
Problem \eqref{eq:min-vol-update} is equivalent to 
$$\min_{W\geq\varepsilon}\max_{\mu\in\mbbR^r}  \mcalL(W,\mu):=G_1(W,\hat W)+\langle W^\top e-e,\mu\rangle. $$
Since $\mcalL(\cdot,\mu)$ is convex, $\mcalL(W,\mu)\to +\infty$ when $\|W\|\to+\infty$, and $\mcalL(W,\cdot)$ is linear, we have strong duality  \cite[Proposition 4.4.2]{Bertsekas2016}, that is, 
$$
\min_{W\geq\varepsilon}\max_{\mu\in\mbbR^r}  \mcalL(W,\mu)=\max_{\mu\in\mbbR^r} \min_{W\geq\varepsilon} \mcalL(W,\mu).
$$
On the other hand, as $W\mapsto \mcalL(W,\mu)$ is separable with respect to each  $W_{jk}$ of $W$, minimizing this function over $W\geq \varepsilon$ reduces to minimizing scalar strongly convex functions of $W_{jk}$ over $W_{jk}\geq \varepsilon$, for $j=1,\ldots,n$, $k=1,\ldots,r$:
\begin{equation*}
\begin{split}
    \min_{W_{jk}\geq \varepsilon} & \Bigg\{\sum_{i=1}^n\frac{(H^\top)_{ik}\hat W_{jk}}{\tilde{v}_{i}} X_{ji}\log\left(\frac{1}{W_{jk}}\right)+ \sum_{i=1}^n (H^\top)_{ik} W_{jk} \\  & \qquad +\lambda_1\big( A_{jk} W_{jk} + \frac12 L_{\phi_1}(\hat W) (W_{jk}-\hat W_{jk})^2\big) + W_{jk}\mu_k\Bigg\}, 
    \end{split}
\end{equation*}
where $\tilde v=H^\top \hat W_{j:}^\top $. This optimization problem can be rewritten as 
$$ \min_{W_{jk}\geq \varepsilon} -b_1 \log( W_{jk}) +b_2 W_{jk}+\frac12 \lambda_1  L_{\phi_1}(\hat W) W_{jk}^2, 
$$
which has the optimal solution 
$$W_{jk}(\mu_k)=\max\left(\varepsilon,\frac12\left(-b_2 + \big(b_2^2 +b_3 b_1 \big)^{1/2}\right)\right),
$$
where 
\begin{align*}
b_1&=\sum_{i=1}^n\frac{(H^\top)_{ik} X_{ji}}{\tilde{v}_{i}}  \hat W_{jk}, \quad b_3= 4 \lambda_1  L_{\phi_1}(\hat W), \\
b_2(\mu_k)&=\sum_{i=1}^n (H^\top)_{ik}  + \lambda_1 \big(A_{jk} - L_{\phi_1}(\hat W) \hat W_{jk}  + \mu_k\big).
\end{align*}
In matrix form, we have \eqref{iMU_Wupdate}. 
We need to find $\mu_k$ such that $\sum_{j=1}^n  W_{jk}(\mu_k)=1$. We have 
$W_{jk}(\mu_k)=\max(\varepsilon, \psi_{jk}(\mu_k)) $, where $\psi_{jk}(\mu_k)=\frac12 (-b_2 + \sqrt{b_2^2 + 4b_3 b_1})$. 
Note that $\mu_k\mapsto W_{jk}(\mu_k)$ is a decreasing function since 
\[
\psi_{jk}'(\mu_k)=\frac12\left(-\lambda_1+\frac{\lambda_1 b_2}{\sqrt{b_2^2 + 4b_3 b_1}}\right) < 0.
\] 
We then apply the bisection method to find the solution of  $\sum_{j=1}^n  W_{jk}(\mu_k)=1$. To determine the segment containing $\mu_k$, we note that if $\varepsilon<1/n$ then $W_{jk}(\tilde\mu_{jk})=1/n$, where $\tilde\mu_{jk}$ is defined in \eqref{mujk}.
Hence, $\mu_k\in [\underline{\mu}_k,\overline{\mu}_k]$, where $\underline{\mu}_k$ and  $\overline{\mu}_k$ are defined in \eqref{mujk}.

\section{BMMe for solving constrained and regularized KL-NMF~\eqref{u_KLNMF}} 
\label{sec:MUe-KL}

Algorithm~\ref{alg:MUe-KL} is BMMe for the specific case of constrained and regularized KL-NMF. 
\begin{algorithm}[ht!]
\caption{BMMe for solving constrained and regularized KL-NMF~\eqref{u_KLNMF}} 
\label{alg:MUe-KL}
\begin{algorithmic}[1]

\STATE Choose initial points $W^{-1}\geq\varepsilon, W^0\geq\varepsilon, H^{-1}\geq\varepsilon, H^0\geq\varepsilon$.
\FOR{$t=1,\ldots$}

\STATE Compute extrapolation points: 
$$
\begin{array}{ll}
 \hatW^t&=W^t + \alpha_W^t[W^t-W^{t-1}]_+, \\
 \hatH^t&=H^t + \alpha_H^t[H^t-H^{t-1}]_+, 
 \end{array}
 $$
\ngi{where $\alpha_W^t$ and $\alpha_H^t$ satisfy the condition of Theorem~\ref{thrm:MUe-beta}.} 

\STATE Update the two blocks of variables: 
\begin{equation} \label{MUe}
\begin{array}{ll}
W^{t+1} \in\underset{W\in\bar\Omega_W}{\argmin}G^t_1(W,\hatW^t), \\
H^{t+1} \in\underset{H\in\bar\Omega_H}{\argmin} G^t_2(H,\hatH^t), 
\end{array}
\end{equation}
where $G_1^t=G_1^{(H^{t})}$ and $G^t_2=G_2^{(W^{t+1})}$ are the majorizers defined in \eqref{eq:W_majorizer} with $H=H^{t}$ and \eqref{eq:H_majorizer} with $W=W^{t+1}$, respectively. 
\ENDFOR
\end{algorithmic}
\end{algorithm}

\bibliographystyle{spmpsci}  
\bibliography{biblio_2}

\section{Supplementary material}

\subsection{$\beta$-NMF with $\beta = 3/2$ for hyperspectral images}   \label{sec:experHSI}

We consider $\beta$-NMF with $\beta = 3/2$ which has been shown to perform well for hyperspectral images; 
see~\cite{fevotte2014nonlinear}.  
We use 4 widely used data sets summarized in Table~\ref{tab:hsidatasets}; see \url{http://lesun.weebly.com/hyperspectral-data-set.html} and  \cite{zhu2017hyperspectral}. 
\begin{table}[h!]
    \centering
    \begin{tabular}{c||c|c|c}
    Data set & $m$ & $n$ & $r$    \\ \hline 
 Urban  & 162 & $307 \times 307$ & $6$     \\
San Diego airport & 158 & $400\times 400$ & $8$    \\
 Pines  & 142 & $145 \times 145$   & $16$     \\
 Cuprite & 188 & $250 \times 191$   & $20$   
    \end{tabular}
    \caption{Summary of the hyperspectral image data sets.} 
    \label{tab:hsidatasets}
\end{table}

A hyperspectral image (HSI) provides a spectral signature for each pixel of the image. The spectral signature measures the fraction of light reflected depending on the wavelength, and HSIs typically measure between 100 and 200 wavelengths. Given such an image, blind hyperspectral unmixing aims to decompose the image into pure materials and abundance maps (which indicates which pixel contains which material and in which proportion). One of the most successful model to perform this task is NMF applied on the wavelength-by-pixel matrix; see \cite{Bioucas-Dias, Maetal2014} and the references therein for more details. 

\begin{figure}[ht!]
\begin{center}
\begin{tabular}{cc}
Urban & San Diego \\  
\includegraphics[width=0.45\textwidth]{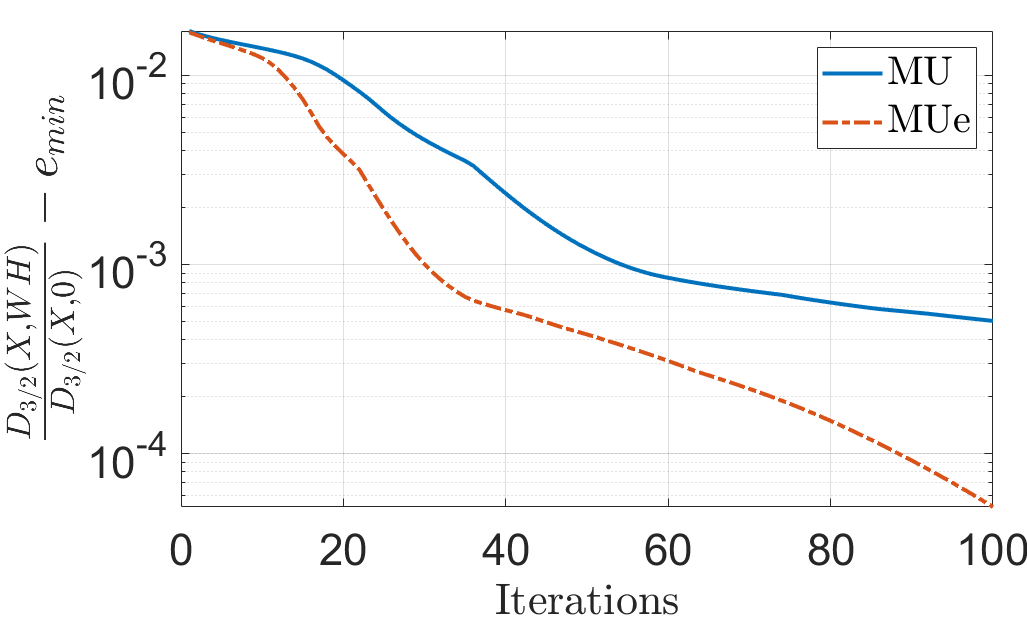} &   
\includegraphics[width=0.45\textwidth]{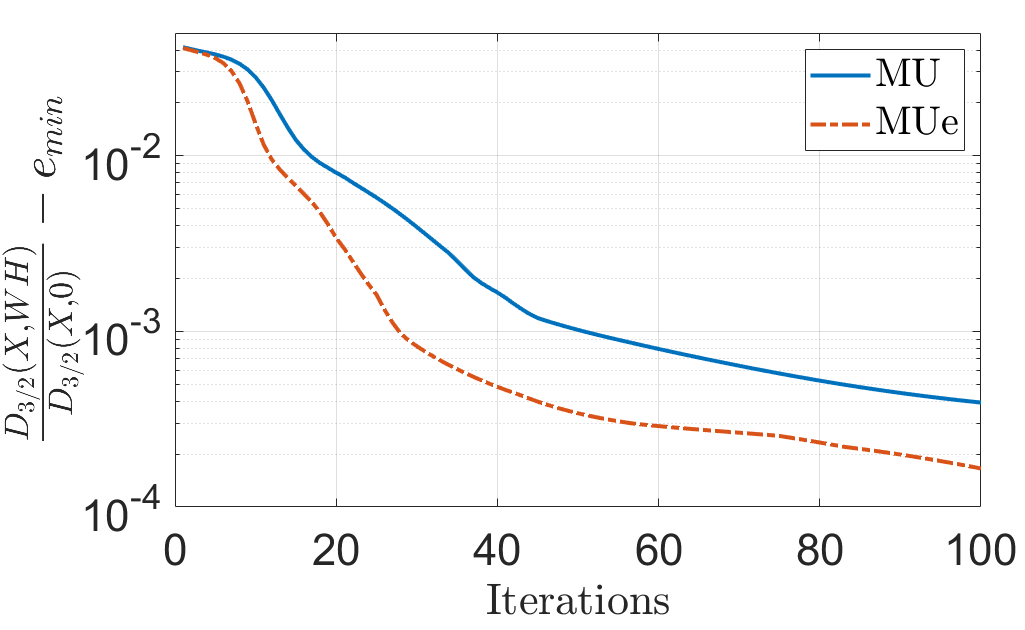} \\
 &   \\ 
Pines & Cuprite  \\ 
\includegraphics[width=0.45\textwidth]{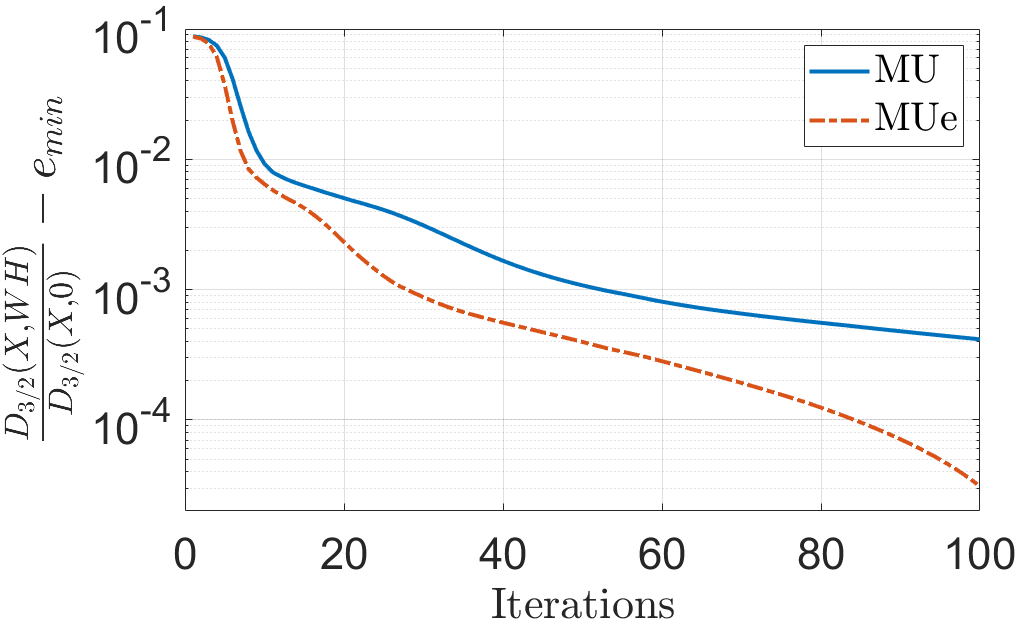} &   
\includegraphics[width=0.45\textwidth]{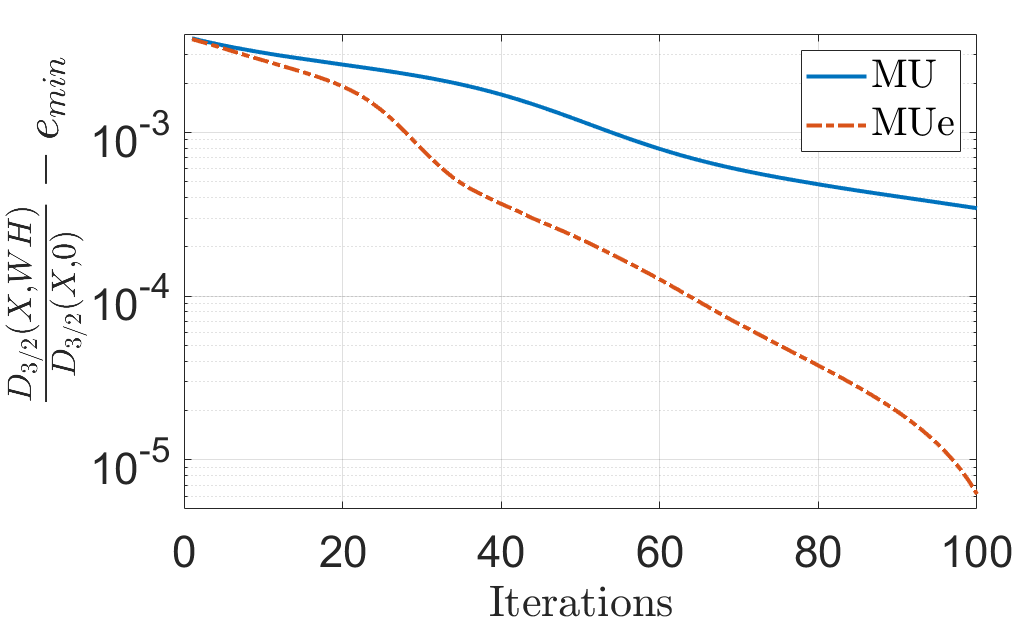} 
\end{tabular}
\caption{Evolution of the objective function of $\beta$-NMF for $\beta = 3/2$ minus the smallest objective  function found among the 10 runs of the two algorithms (denoted $e_{min}$): MU vs.\ MUe on the four hyperspectral data sets ; see Table~\ref{tab:hsidatasets}. 
We report the median over 10 randomly generated initial matrices for both algorithms.} 
\label{fig:hsimages}
\end{center}
\end{figure}

Figure~\ref{fig:hsimages} displays the median relative objective function values minus the best solution found among 10 random initializations. 
Table~\ref{tab:iterMUvsMUe} provides the minimum, median and maximum number of iterations for MUe to obtain an objective function value smaller than MU with 100 iterations, for the 10 random initializations. 
On average, MUe requires less than 50 iterations, meaning that MUe is more than twice faster than MU. 
MUe provides a significant acceleration over MU for all data sets and all initializations; in the worst case (for a total of 40 runs: 4 data sets with 10 random initializations each), it takes MUe 55 iterations to reach the error of MU with 100 iterations. 
\begin{center}
\begin{table}[h!]
\begin{center}  
\begin{tabular}{c||c|c|c} 
Data set & min & meadian & max \\ \hline 
San Diego  &  42  & 46   & 52   \\ 
Urban      &  37  & 43   & 55     \\ 
Cuprite    &  39  & 40.5 & 42   \\ 
Pines      &  43  & 47   & 49 \\ 
\end{tabular} 
\caption{Number of iterations needed for MUe to obtain an objective function value smaller than MU after 100 iterations. We report the minimum, median and maximum number over the 10 random initializations.}
\label{tab:iterMUvsMUe}
\end{center}
\end{table}
\end{center}

\subsection{KL-NMF for topic modeling and imaging}   \label{sec:addtopic}

As explained in Section~\ref{sec:KLnmf}, KL-NMF is widely used for imaging and topic modeling. 
In this section, we report extensive results for both applications, following the experimental setup of~\cite{HienNicolasKLNMF} (in particular, we use $r=10$ in all cases).  
We compare MU and MUe to the state-of-the-art algorithm CCD~\cite{Hsieh2011}. In the extensive numerical experiments reported in  \cite{HienNicolasKLNMF}, MU and CCD were the best two algorithms for KL-NMF.

\paragraph{Dense facial images} 

\begin{figure}[ht!]
\begin{center}
\begin{tabular}{cc}
CBCL & Frey \\  
\includegraphics[width=0.45\textwidth]{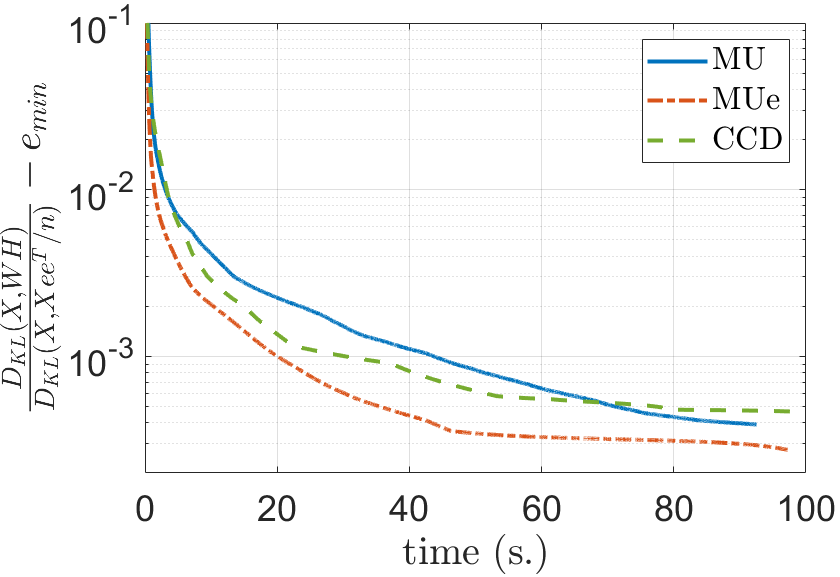} &   
\includegraphics[width=0.45\textwidth]{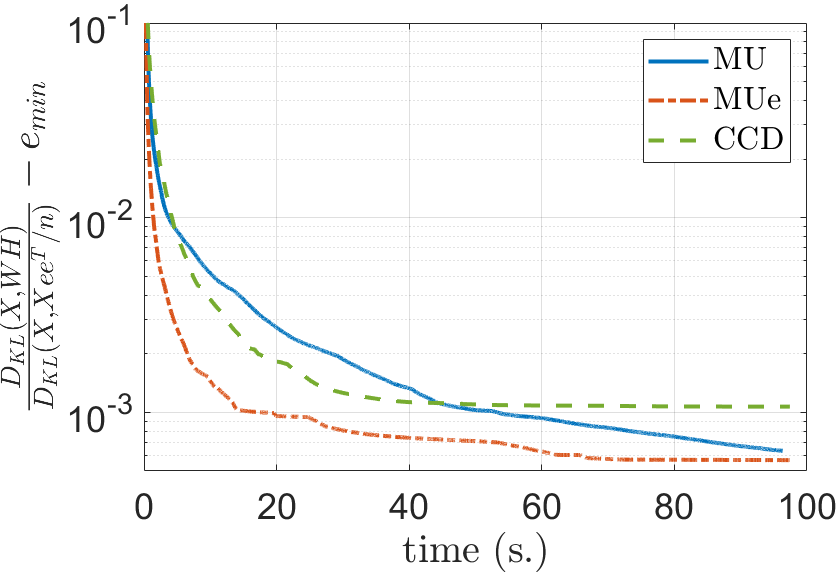} \\
 &   \\ 
ORL & UMIST  \\ 
\includegraphics[width=0.45\textwidth]{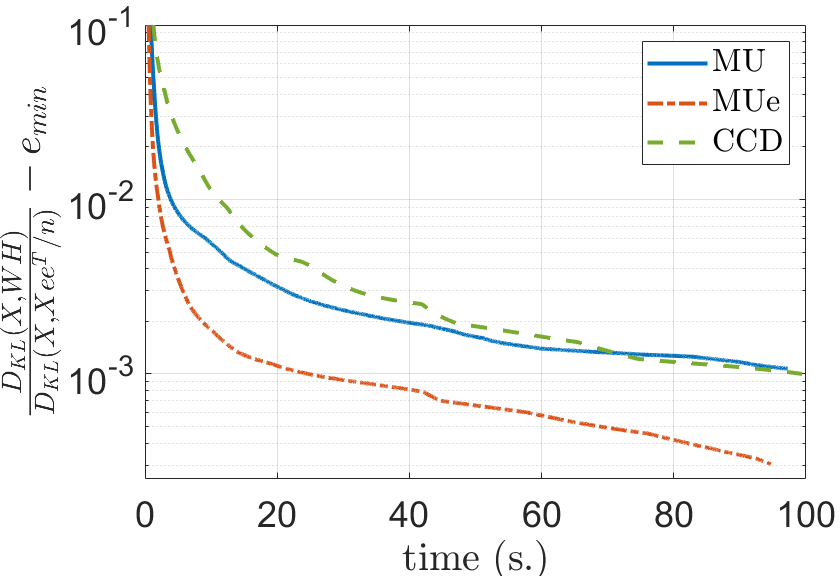} &   
\includegraphics[width=0.45\textwidth]{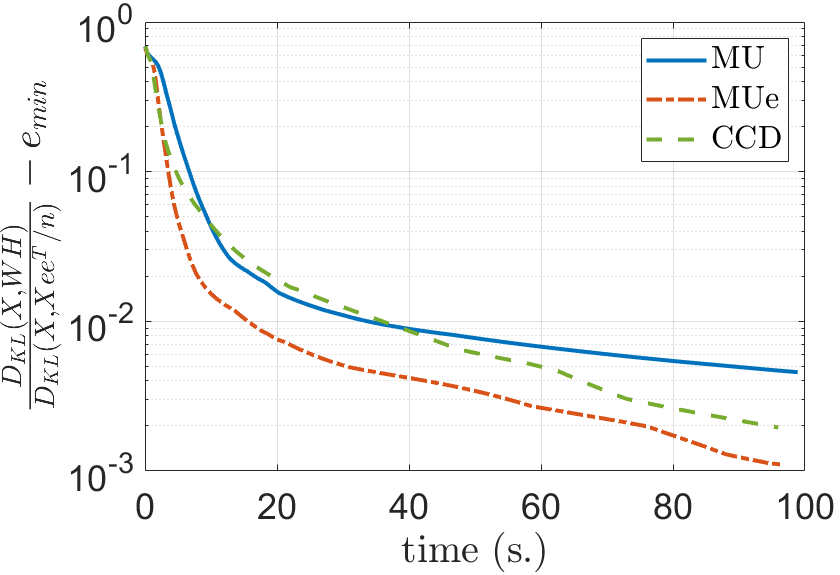} 
\end{tabular}
\caption{Evolution of the objective function of KL-NMF minus the smallest objective  function found among the 10 runs of the three algorithms (denoted $e_{min}$) on the four facial image data sets.  
We report the median over 10 randomly generated initial matrices for the three algorithms. \revision{Note that MU, MUe and CCD respectively perform on average 9842, 9745 and 1548 iterations for CBCL, 13715, 13630 and 1557 iterations for Frey, 
7579, 6949, and 619 for ORL, and 1544, 1530 and 242 for UMIST within 100 seconds.}}
\label{fig:facialimages}
\end{center}
\end{figure}

We use 4 popular facial image summarized in Table~\ref{tab:facedatasets}. 
\begin{table}[h!]
    \centering
    \begin{tabular}{c||c|c}
    Data set & $m$ (\# pixels) & $n$ (\# images)    \\ \hline 
CBCL   & $19 \times 19$ &  2429   \\
Frey &  $28 \times 20$ & 1965     \\ 
ORL   &  $112 \times 92$ &      565 \\
UMist &  $112 \times 92$ &         400
    \end{tabular}
    \caption{Summary of the facial image data sets.} 
    \label{tab:facedatasets}
\end{table}  
Figure~\ref{fig:facialimages} displays the median relative objective function values minus the best solution found among 10 random initializations. 
For these dense data sets, MUe performs better than CCD, which was not the case of MU that performs on average worse than CCD on dense data sets~\cite{HienNicolasKLNMF}. 
This means that not only MUe provides a significant acceleration of MU, but also outperforms the state-of-the-art algorithm CCD for KL NMF on dense data sets.

\paragraph{Sparse document data sets} 

\begin{figure}[ht!]
\begin{center}
\begin{tabular}{cc}
classic & hitech \\  
\includegraphics[width=0.47\textwidth]{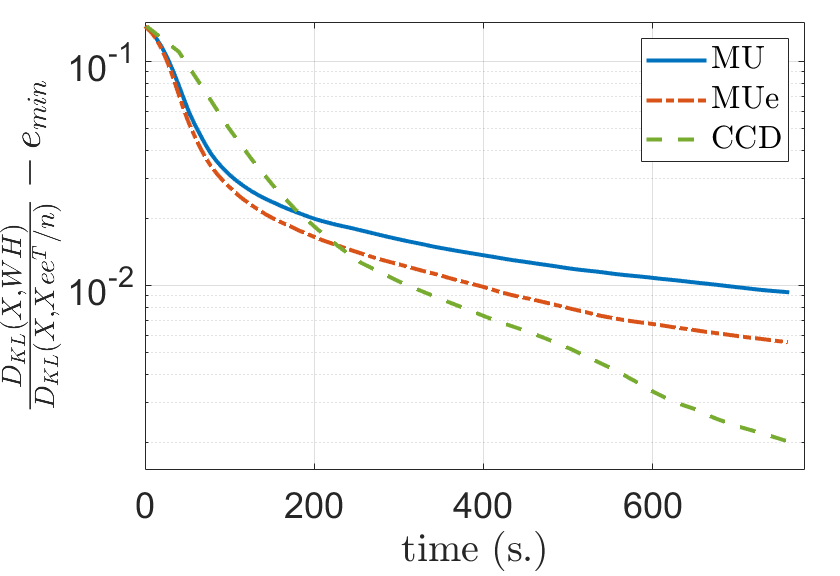} &   
\includegraphics[width=0.47\textwidth]{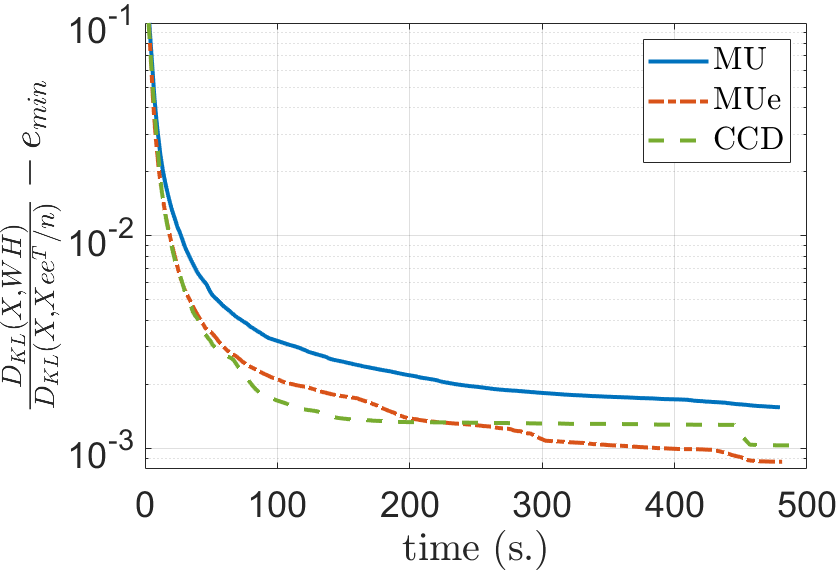} \\
 &   \\ 
la1 & sports  \\ 
\includegraphics[width=0.47\textwidth]{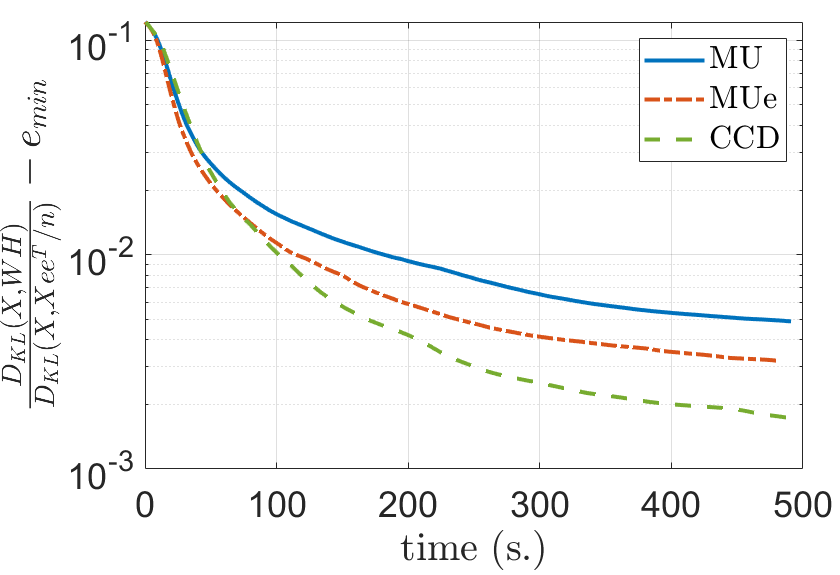} &   
\includegraphics[width=0.47\textwidth]{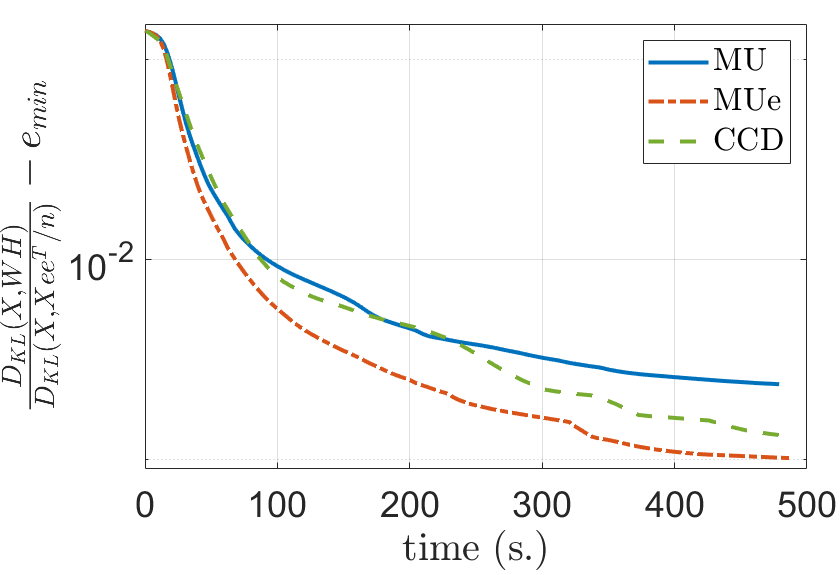} 
\end{tabular}
\caption{Evolution of the objective function of KL-NMF minus the smallest objective  function found among the 10 runs of the three algorithms (denoted $e_{min}$) on the four facial image data sets.  
We report the median over 10 randomly generated initial matrices for the three algorithms. \revision{Note that MU, MUe and CCD respectively perform on average 123, 123 and 33 iterations for classic, and 885, 886 and 343 iterations for hitech, 219, 217, and 65 for la1, and 178, 178 and 56 for sports in the considered time intervals.}}
\label{fig:docdsets}
\end{center}
\end{figure}

We use 4 document data sets summarized in Table~\ref{tab:facedatasets}.  
\begin{table}[h!]
    \centering
    \begin{tabular}{c||c|c|c}
    Data set & $m$ (\# documents) & $n$ (\# words)  & sparsity (\% zeros)   \\ \hline 
classic   & 7094 &  41681   & 99.92 \\
hitech   & 2301  &   10080  &  98.57 \\
la1 &  3204  &  31472   &  99.52 \\ 
sports &  8580 & 14870  &   99.14      
    \end{tabular}
    \caption{Summary of the document data sets; see~\cite{zhong2005generative} for more details.} 
    \label{tab:docdatasets}
\end{table} 
Figure~\ref{fig:docdsets} displays the median relative objective function values minus the best solution found among 10 random initializations. 
For sparse data sets, MUe outperforms MU, as in all our other experiments so far. 
However, MUe does not outperform CCD which performs better on two data sets (classic, la1). 
Hence, for sparse data sets, CCD is competitive with MUe, although no algorithm seem to outperform the other one.

\subsection{Min-vol KL-NMF for audio data sets}   \label{sec:addexperaudio}

NMF has been used successfully to perform blind audio source separation. The input matrix is a spectrogram that records the activations of the frequencies over time. Applying NMF on this spectrogram allows us to recover the frequency response of the sources as the columns of $W$, and the activations of the source over time as the rows of $H$~\cite{smaragdis2003non}. 
For example, applying NMF on a piano recording will extract automatically the spectrum of the notes (their harmonics) and their activation over time, so that NMF can for example be used for automatic music transcription~\cite{benetos2019automatic}.

 In this section, we report the results obtained for the same algorithms as in Section~\ref{sec:experminvolKL} on the audio data sets "Mary had a little lamb" and the audio sample from  \cite{10.1162/neco.2008.04-08-771}, see Table~\ref{tab:audiodatasets}. 
\begin{table}[h!]
    \centering
    \begin{tabular}{c|c|c|c|c}
    Data set & $m$ & $n$ & $r$ & $\tilde{\lambda}$   \\ \hline 
 Mary had a little lamb  & 129 & $586$ & $4$ &  0.3   \\
Prelude from J.S.-Bach & 129 & $2582$ & $16$  & 0.04 \\
   \revision{Sample from} \cite{10.1162/neco.2008.04-08-771}  & $513$ & $676$   & $7$  & 0.015  \\
    \end{tabular}
    \caption{Audio data sets.} 
    \label{tab:audiodatasets}
\end{table} 
The parameter $\tilde{\lambda}$ is chosen as the initial ratio between $\lambda_1\log\det(W^\top W+\delta I)$ and $D_{KL} (X,WH)$. In practice, given the initial point $(W^{(0)}, H^{(0)})$, we set $\tilde{\lambda}$, and the min-vol weight parameter $\lambda_1$ is determined using the formula: 
\[
\lambda_1 = \tilde{\lambda} \frac{D_{KL} (X,W^{(0)}H^{(0)})}{|\log\det(W^{(0),\top} W^{(0)}+\delta I)|}.
\]
The value for $\tilde{\lambda}$ for each dataset is extracted from \cite{leplat2020blind}, where successful audio source separation with min-vol $\beta$-NMF was illustrated.

\begin{figure}[ht!]
\begin{center}
\begin{tabular}{cc}
\quad Mary had a little lamb & \qquad  Sample from \cite{10.1162/neco.2008.04-08-771} \\  
\includegraphics[width=0.47\textwidth]{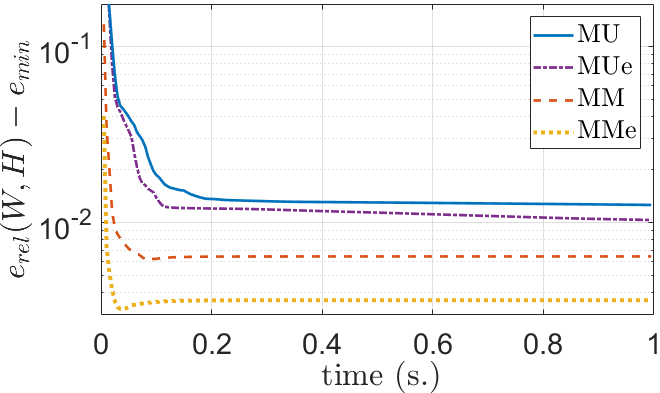} &   
\includegraphics[width=0.47\textwidth]{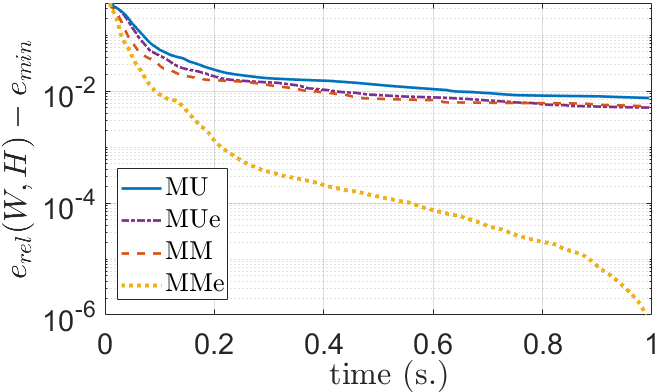} \\
 &   \\ 
\end{tabular}
\caption{Benchmarked algorithms on first and third data set detailed in Table~\ref{tab:audiodatasets}. Evolution of the median relative errors from Equation~\cref{eq:rel_error_minvolklnmf} for 20 random initial initializations minus the smallest relative error found (denoted $e_{min}$). \revision{Note that MU, MUe, MM and MMe respectively perform on average 552, 590, 2468 and 2455 iterations for Mary had a little lamb, and 151, 151, 253 and 246 iterations for the Sample from \cite{10.1162/neco.2008.04-08-771} within one second.}}
\label{fig:mary_Bertin}
\end{center}
\end{figure}

Figure \ref{fig:mary_Bertin} shows the median relative errors obtained by each algorithm for these two datasets across 20 runs, each result normalized by subtracting the smallest relative error achieved among all runs. 
The observations are similar as for the data set presented in the paper, namely MMe outperforms the other algorithms, showing once again that our BMMe framework provides a significant acceleration effect. Moreover, MUe allows us to accelerate MU.

\end{document}